%% file: 0_main.tex
\documentclass[sigconf]{acmart}
\usepackage{booktabs}
\usepackage{multirow}
\newtheorem{myDef}{Definition}
\usepackage[ruled,linesnumbered]{algorithm2e}
\usepackage{algpseudocode}
\usepackage{subfigure}
\usepackage{xcolor}
\usepackage[toc,title]{appendix}
\usepackage{enumitem}
\newcommand{\modelname}{HyperIMBA}

\AtBeginDocument{%
  \providecommand\BibTeX{{%
    \normalfont B\kern-0.5em{\scshape i\kern-0.25em b}\kern-0.8em\TeX}}}

\begin{document}

\copyrightyear{2023} 
\acmYear{2023} 
\setcopyright{acmlicensed}\acmConference[WWW '23]{Proceedings of the ACM Web Conference 2023}{May 1--5, 2023}{Austin, TX, USA}
\acmBooktitle{Proceedings of the ACM Web Conference 2023 (WWW '23), May 1--5, 2023, Austin, TX, USA}
\acmPrice{15.00}
\acmDOI{10.1145/3543507.3583403}
\acmISBN{978-1-4503-9416-1/23/04}

\title{Hyperbolic Geometric Graph Representation Learning for Hierarchy-imbalance Node Classification}


\author{Xingcheng Fu}
\orcid{0000-0002-4643-8126}
\affiliation{%
  \institution{School of Computer Science and Engineering, BDBC,\\ Beihang University}
  \city{Beijing}
  \country{China}
}
\email{fuxc@act.buaa.edu.cn}

\author{Yuecen Wei}
\orcid{0000-0002-0218-948X}
\affiliation{%
  \institution{Guangxi Key Lab of Multi-source Information Mining Security, Guangxi Normal University}
  \city{Guilin}
  \country{China}
}
\email{weiyc@stu.gxnu.edu.cn}

\author{Qingyun Sun}
\orcid{0000-0003-1930-3848}
\affiliation{%
  \institution{School of Computer Science and Engineering, BDBC,\\ Beihang University}
  \city{Beijing}
  \country{China}
}
\email{sunqy@act.buaa.edu.cn}

\author{Haonan Yuan}
\orcid{0000-0001-9205-8610}
\affiliation{%
  \institution{School of Computer Science and Engineering, BDBC,\\ Beihang University}
  \city{Beijing}
  \country{China}
}
\email{yuanhn@act.buaa.edu.cn}

\author{Jia Wu}
\orcid{0000-0002-1371-5801}
\affiliation{%
  \institution{School of Computing,\\ Macquarie University}
  \city{Sydney}
  \country{Australia}
}
\email{jia.wu@mq.edu.au}

\author{Hao Peng}
\orcid{0000-0001-7422-630X}
\affiliation{%
  \institution{School of Cyber Science and Technology, BDBC,\\ Beihang University}
  \city{Beijing}
  \country{China}
}
\email{penghao@act.buaa.edu.cn}

\author{Jianxin Li}
\orcid{0000-0001-5152-0055}
\affiliation{%
  \institution{School of Computer Science and Engineering, Beihang University, Zhongguancun Lab}
  \city{Beijing}
  \country{China}
}
\email{lijx@act.buaa.edu.cn}

% \author{Xingcheng Fu{$^{1,2}$}, Yuecen Wei{$^{3,4}$}, Qingyun Sun{$^{1,2}$}, Haonan Yuan{$^{1,2}$}, Jia Wu{$^{5}$}, Hao Peng{$^{1}$}, Jianxin Li{$^{1,2}$}}
% \affiliation{
%   \institution{
%   $^1$
%   Beijing Advanced Innovation Center for Big Data and Brain Computing, Beihang University, Beijing 100191, China\\
%   $^2$
%   School of Computer Science and Engineering, Beihang University, Beijing 100191, China\\
%   $^3$
%   Guangxi Key Lab of Multi-source Information Mining \& Security, Guangxi Normal University, Guilin, China\\
%   $^4$
%   School of Computer Science and Engineering, Guangxi Normal University, Guilin, China\\
%   $^5$
%   School of Computing, Macquarie University, Sydney, Australia\\
%   }
%   \country{}
% }
% \author{
%     Anonymous submission
% }

% \email{{lijx,fuxc,sunqy,jicheng,penghao}@act.buaa.edu.cn, chiachiun_than@buaa.edu.cn, jia.wu@mq.edu.au}

\renewcommand{\shortauthors}{Xingcheng Fu, et al.}

\begin{abstract}
Learning unbiased node representations for imbalanced samples in the graph has become a more remarkable and important topic. 
For the graph, a significant challenge is that the topological properties of the nodes (e.g., locations, roles) are unbalanced (topology-imbalance), other than the number of training labeled nodes (quantity-imbalance). 
Existing studies on topology-imbalance focus on the location or the local neighborhood structure of nodes, ignoring the global underlying hierarchical properties of the graph, i.e., hierarchy. 
In the real-world scenario, the hierarchical structure of graph data reveals important topological properties of graphs and is relevant to a wide range of applications. 
We find that training labeled nodes with different hierarchical properties have a significant impact on the node classification tasks and confirm it in our experiments. 
It is well known that hyperbolic geometry has a unique advantage in representing the hierarchical structure of graphs. 
Therefore, we attempt to explore the hierarchy-imbalance issue for node classification of graph neural networks with a novelty perspective of hyperbolic geometry, including its characteristics and causes. 
Then, we propose a novel hyperbolic geometric hierarchy-imbalance learning framework, named \modelname, to alleviate the hierarchy-imbalance issue caused by uneven hierarchy-levels and cross-hierarchy connectivity patterns of labeled nodes.
Extensive experimental results demonstrate the superior effectiveness of \modelname~for hierarchy-imbalance node classification tasks. 
\end{abstract}

% <concept_id>10010147.10010257.10010293.10010294</concept_id>
% <concept_desc>Computing methodologies~Neural networks</concept_desc>
% <concept_id>10010147.10010257.10010293.10010319</concept_id>
% <concept_desc>Computing methodologies~Learning latent representations</concept_desc>
% <concept_id>10002950.10003624.10003633.10003643</concept_id>
% <concept_desc>Mathematics of computing~Graphs and surfaces</concept_desc>

% \ccsdesc[500]{Computing methodologies~Neural networks}
% \ccsdesc[500]{Computing methodologies~Learning latent representations}
% \ccsdesc[500]{Mathematics of computing~Graphs and surfaces}


\maketitle

\input{1_introduction}

\input{3_preliminary}

\input{4_model}

\input{5_experiment}

\input{6_conclusion}

\begin{acks}
The corresponding author is Jianxin Li. 
The authors of this paper were supported by the NSFC through grants (No.U20B2053), and the Australian Research Council (ARC) Projects Nos. DE200100964, LP210301259, and DP230100899. 
\end{acks}

\bibliographystyle{ACM-Reference-Format}
\bibliography{reference}

\end{document}

%% file: 1_introduction.tex
\section{Introduction}
In recent years, graph representation learning has shown its effectiveness in capturing the irregular but related complex structures in graph data~\cite{hamilton2017inductive, zhang2020deep,li2022curvature,fu2021ace,sun2021sugar}. 
% The core assumption of graph representation learning is that topological properties are critical to representational capability. 
With the intensive studies and wide applications of graphs~\cite{nastase2015survey,sun2022self,li2023adaptive,yu2022cross}, some recent works~\cite{NickelK17Poincare,HGCN_ChamiYRL19,topping2021understanding} show that the geometric properties of graph topology play a crucial role in graph representation learning. 
Among the variety of topological properties, the hierarchy is a ubiquitous and significant property of graphs. 
In this work, we focus on the semi-supervised unbalanced node classification task for a graph with hierarchy.

\begin{figure*}[!t]
\centering
\subfigure[Quantity-imbalance.]{
\includegraphics[width=0.22\linewidth]{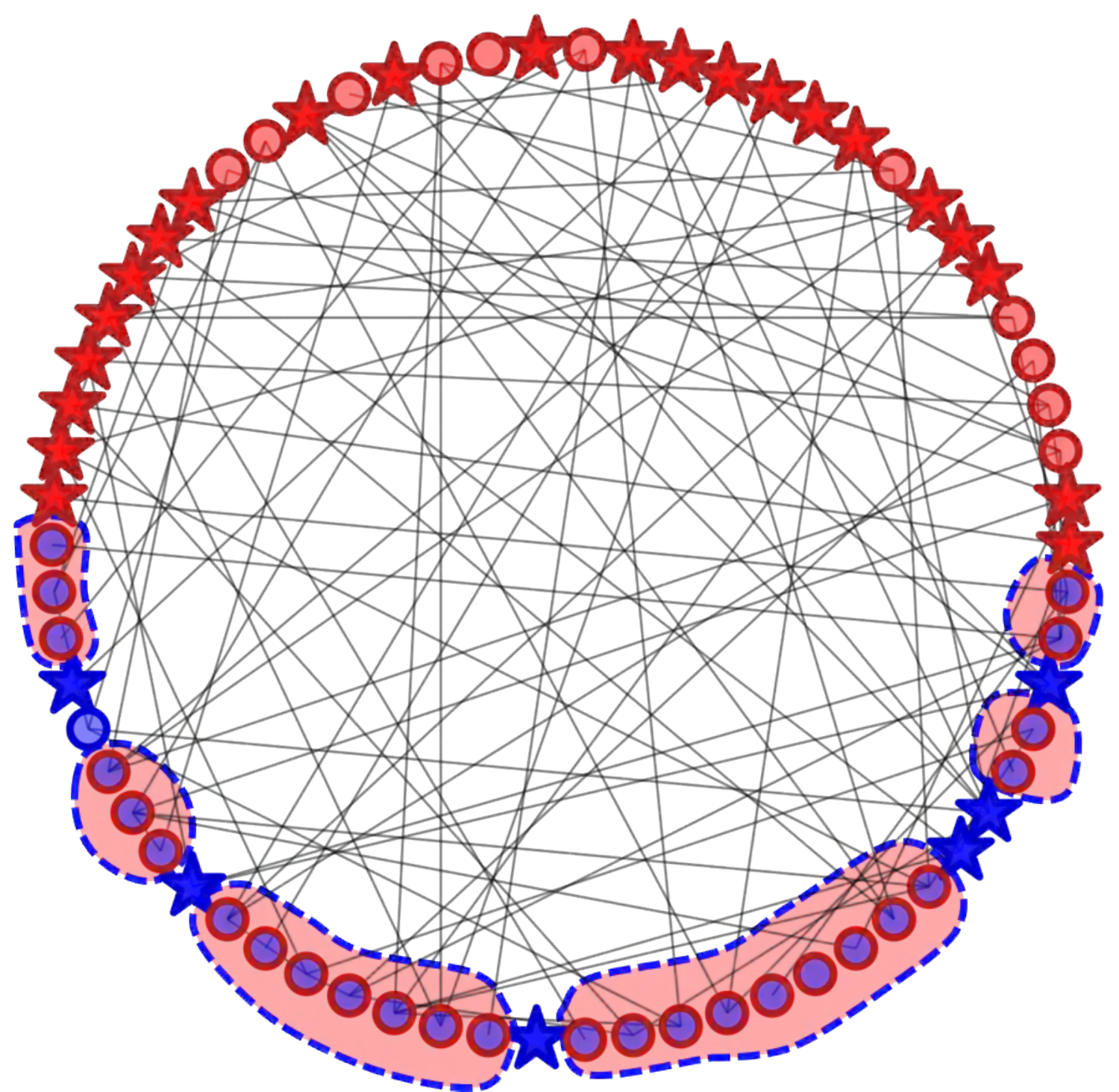}
%\caption{fig1}
}%
\hspace{0.2cm}
\subfigure[Position-imbalance.]{
\includegraphics[width=0.22\linewidth]{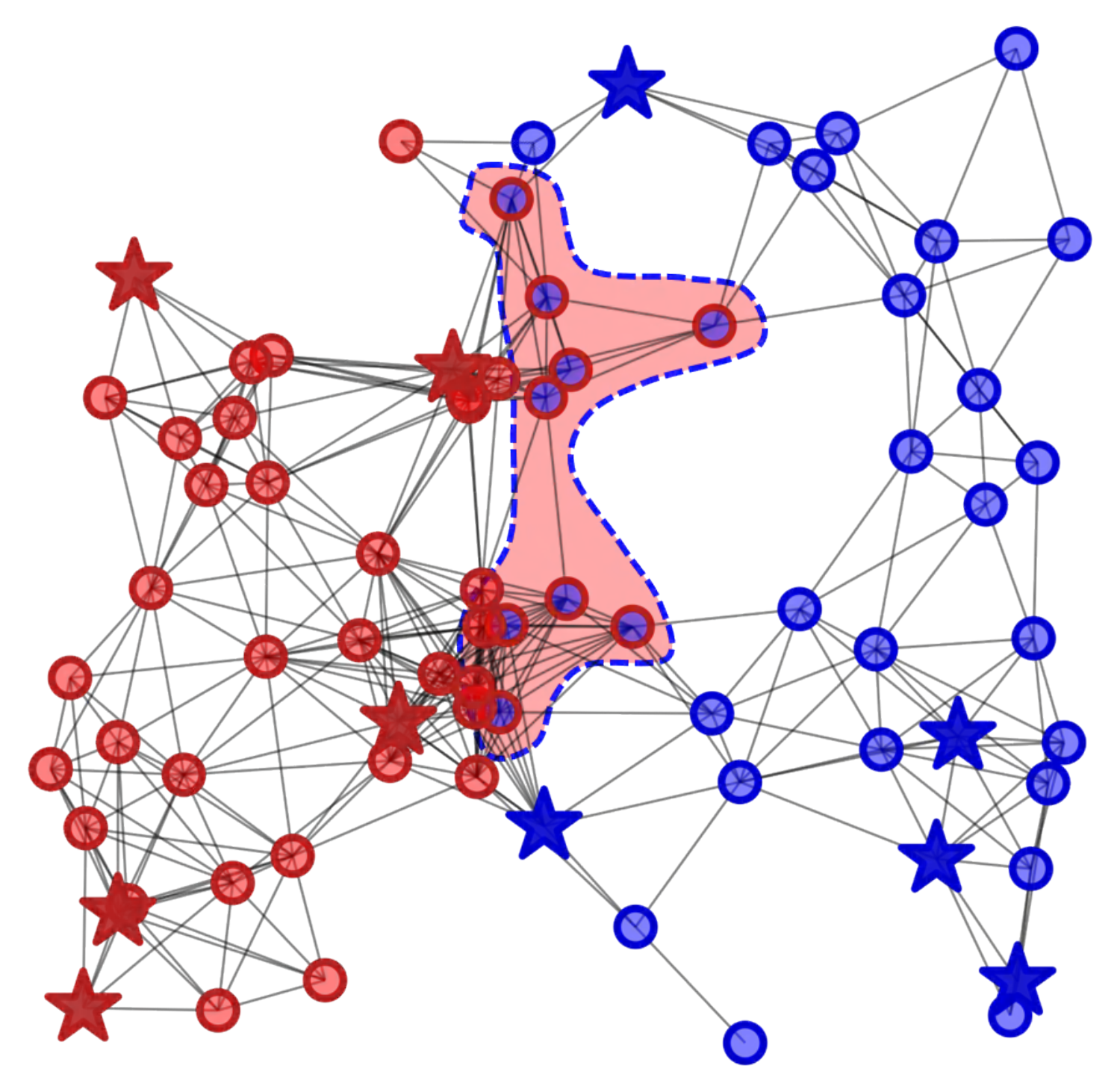}
%\caption{fig2}
}%
\hspace{0.2cm}
\subfigure[Hierarchy-imbalance.]{
\includegraphics[width=0.48\linewidth]{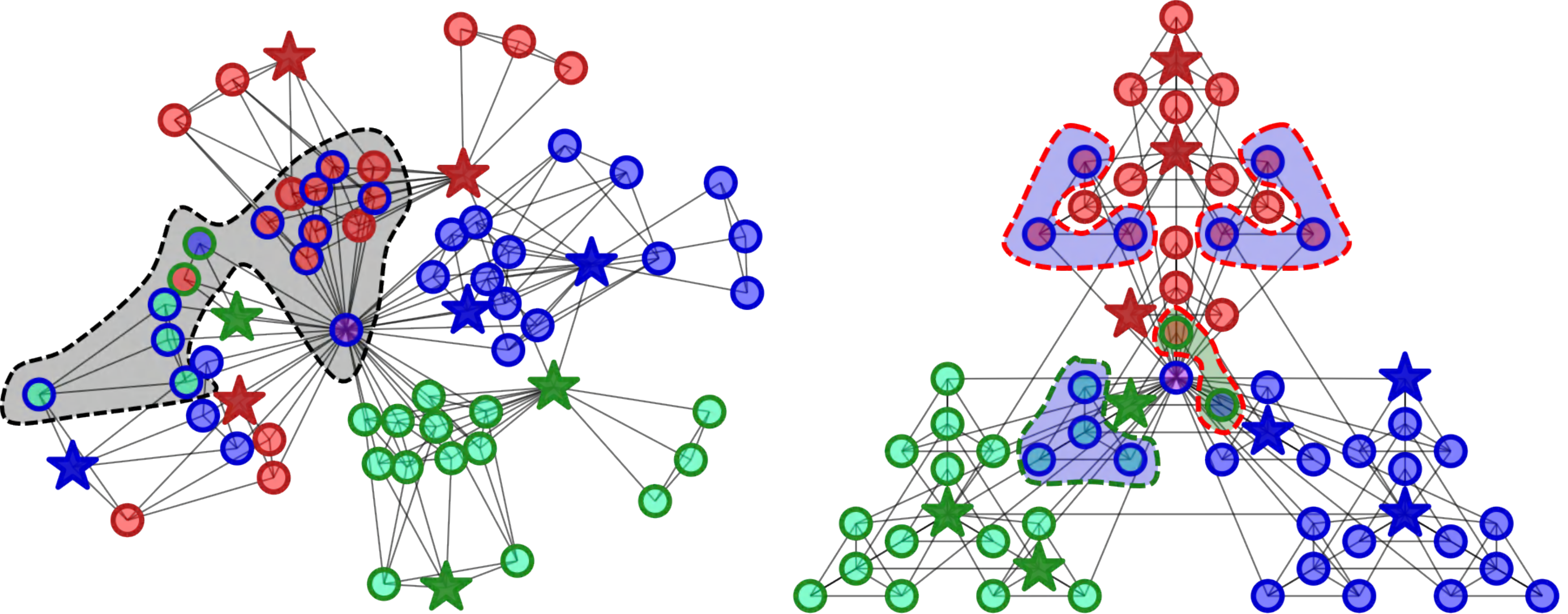}
%\caption{fig2}
}%
\centering
\vspace{-1em}
\caption{Representative cases of imbalanced node classification on the graphs by label propagation. Colors represents different types of nodes, where the fill colors of nodes represent the ground truth and the border colors represent the classification result by label propagation. Incorrect results are marked with the dashed regions. The star and circular symbols represent the labeled nodes and unlabeled nodes, respectively. }
\vspace{-1em}
\label{fig:example}
\end{figure*}

Due to the cost of labeling in the real-world, the number of labels in different classes is always imbalanced. 
Most of the existing studies on imbalanced learning of graphs focus on the imbalanced number of labeled nodes in different classes, i.e., quantity-imbalance~\cite{sun2009classification,he2009learning,haixiang2017learning,lin2017focal,park2021graphens,wang2021distance}, and a toy example as shown in Figure~\ref{fig:example}~(a).
With the in-depth study of the topological properties and the learning mechanism of GNNs, some recent works have focused on the problem of an unbalanced distribution of labeled nodes in topological positions, i.e., topological position imbalance.
The (topological) position-imbalance~\cite{chen2021topology,song2022tam,pastel2022} is caused by the uneven location distribution of labeled nodes in the topological space. 
% For example, we may only have the label information of some certain local community in a social network, which leads to a serious imbalance in the location distribution of labeled nodes.
As shown in Figure~\ref{fig:example}~(b), since the graph neural networks~\cite{GCN,GAT,hamilton2017inductive} learn node representations in the message-passing paradigm~\cite{gilmer2017neural}, even if the labeled nodes have balanced quantity, the uneven position distribution of nodes leads to low-quality of label propagation.
% ReNode~\cite{chen2021topology}, provides an understanding of the topology-imbalance issue  from the perspective of label propagation and proposes a sample re-weighting method. 
% TAM~\cite{song2022tam} hypothesize that the false positive cases is highly affected by the neighbor label distribution of each node, and adaptively adjusts the margin of objective accordingly based on node topology compared to class statistics.
However, the existing works~\cite{chen2021topology,song2022tam, pastel2022} only focus on the position and neighborhood information of nodes in the topology structure, and they are difficult to handle the label imbalance issue caused by implicit topological properties of graphs. 
To sum up, \textit{the imbalance issue exploring the topological properties is still in its infancy. }

\textbf{Hierarchy-imbalance issue.} 
Hierarchy is an important topological property of graphs, which simultaneously reproduces the unique properties of scale-free topology and the high clustering of nodes~\cite{ravasz2003hierarchical,vazquez2002large} and can be widely observed in nature, from biology to language to some social networks~\cite{ravasz2003hierarchical,albert1999diameter}, and reflects the important role of nodes in the network. 
Since graphs with the hierarchical structures are scale-free (with an exponential number growth of nodes and power-law degree distribution)~\cite{barabasi1999emergence} and highly modularity (with high connectivity)~\cite{dorogovtsev2002pseudofractal}, it is difficult to measure label imbalance on hierarchical graphs simply by the quantity and location of nodes. 
For example, the graph in Figure~\ref{fig:example} (c) with a hierarchical structure has a balanced quantity (with the same number of labeled nodes in each class) and balanced position (with the same shortest distance to the graph center and degree distribution) of labeled nodes. However, the blue- and green-class labeled nodes occupy higher hierarchy-level roles, resulting in a large number of errors in the classification results. 
It can be observed that the distribution of labeled nodes in the hierarchy can seriously affect the decision boundary shift of the classifier. 
Compared with quantity- and position-imbalance issues, exploring the hierarchy-imbalance issue has two major challenges: 
\textbf{(1) Implicit topology: }hierarchy-imbalance is caused by the uneven distribution of labeled nodes in implicit topological properties, which is difficult to measure intuitively. 
\textbf{(2) Hierarchical connectivity: }hierarchy of the graph introduces more complex connectivity patterns for nodes, which are difficult to be directly observed and quantified in a single way. 
Therefore, a natural problem is, "\textit{How to effectively and efficiently measure the hierarchy for each node of graphs, and how does hierarchy-imbalanced label information affect the classification results by the message-passing mechanism?} "

\textbf{Present work.}
To solve the above problem, we first give a quantitative analysis for understanding and explore the hierarchy-imbalance issue. 
Furthermore, inspired by the success of hyperbolic graph learning~\cite{NickelK17Poincare,PoincareGlove,HGCN_ChamiYRL19} for hierarchy-preserving~\cite{Krioukov2010Hyperbolic,papadopoulos2012popularity}, we propose an effective metric to measure the hierarchy of labeled nodes using hyperbolic geometric embedding.  
Then we propose a novel \underline{\textbf{Hyper}}bolic Geometric Hierarchy-\underline{\textbf{IMBA}}lance Learning (\textbf{\modelname}) framework to re-weight the label information propagation and adjust the objective margin accordingly based on the node hierarchy. 
The key insight of \modelname~is to use the graph geometric method to deal with the imbalance issue caused by the topological geometric properties.
Specifically, based on the Poincar\'e model, we design a novel \textbf{Hierarchy-Aware Margin (HAM)} to reduce the decision boundary bias caused by hierarchy-imbalance labeled nodes. 
Then we design a \textbf{Hierarchy-aware Message-Passing Neural Network (HMPNN)} mechanism based on the \textit{class-aware Ricci curvature weight}, which measures the influence from the label information and connectivity of neighborhood, alleviating the over-squashing caused by message-passing of cross-hierarchy connectivity pattern by re-weight the "backbone" paths. 
Overall, the contributions are summarized as follows: 

\begin{itemize}[leftmargin=*]
\item 
For the first time, we explore the hierarchy-imbalance node representation learning as a new issue of topological imbalance topic for semi-supervised node classification. 
\item 
We propose a novel training framework, named~\modelname, to alleviate the hierarchy-imbalance issue by designing two key mechanisms: HAM captures the implicit hierarchy of labeled nodes to adjust the decision boundary margins, and HMPNN re-weights the path of supervision information passing according to the cross-hierarchy connectivity pattern.
\item Extensive experiments on synthetic and real-world datasets demonstrate a significant and consistent improvement and provide insightful analysis for the hierarchy-imbalance issue. 
\end{itemize}

%% file: 3_preliminary.tex
\begin{figure*}[ht]
\centering
% \subfigure[Distributions analysis of $\mathrm{BA}_{256}$ and $\mathrm{HN}_{256}$.]{
% \includegraphics[width=0.32\linewidth]{Degree_distribution.pdf}
% %\caption{fig1}
% }%
\subfigure[Quantitative analysis of hierarchical network.]{
\includegraphics[width=0.55\linewidth]{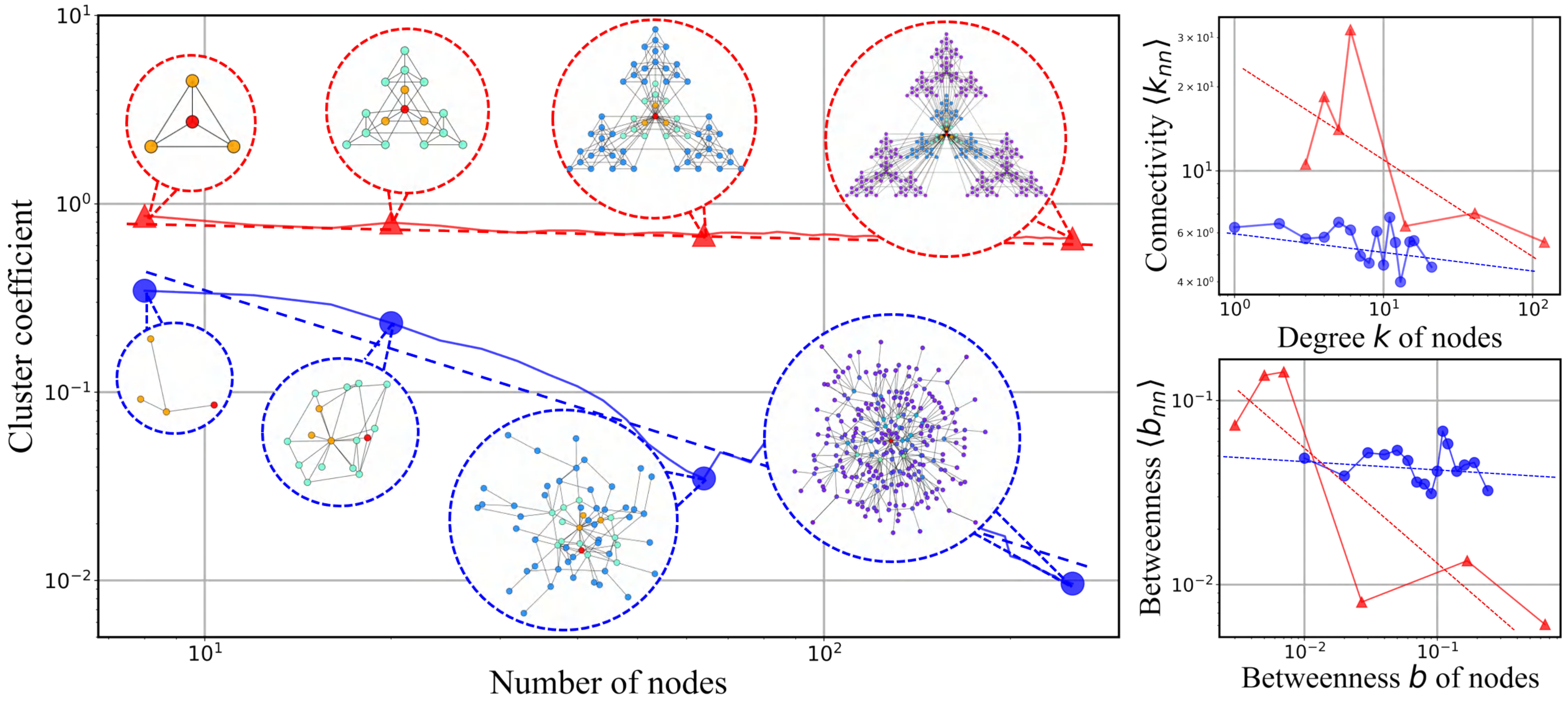}
%\caption{fig2}
}
\subfigure[Tree, hierarchical structure in hyperbolic space.]{
\includegraphics[width=0.41\linewidth]{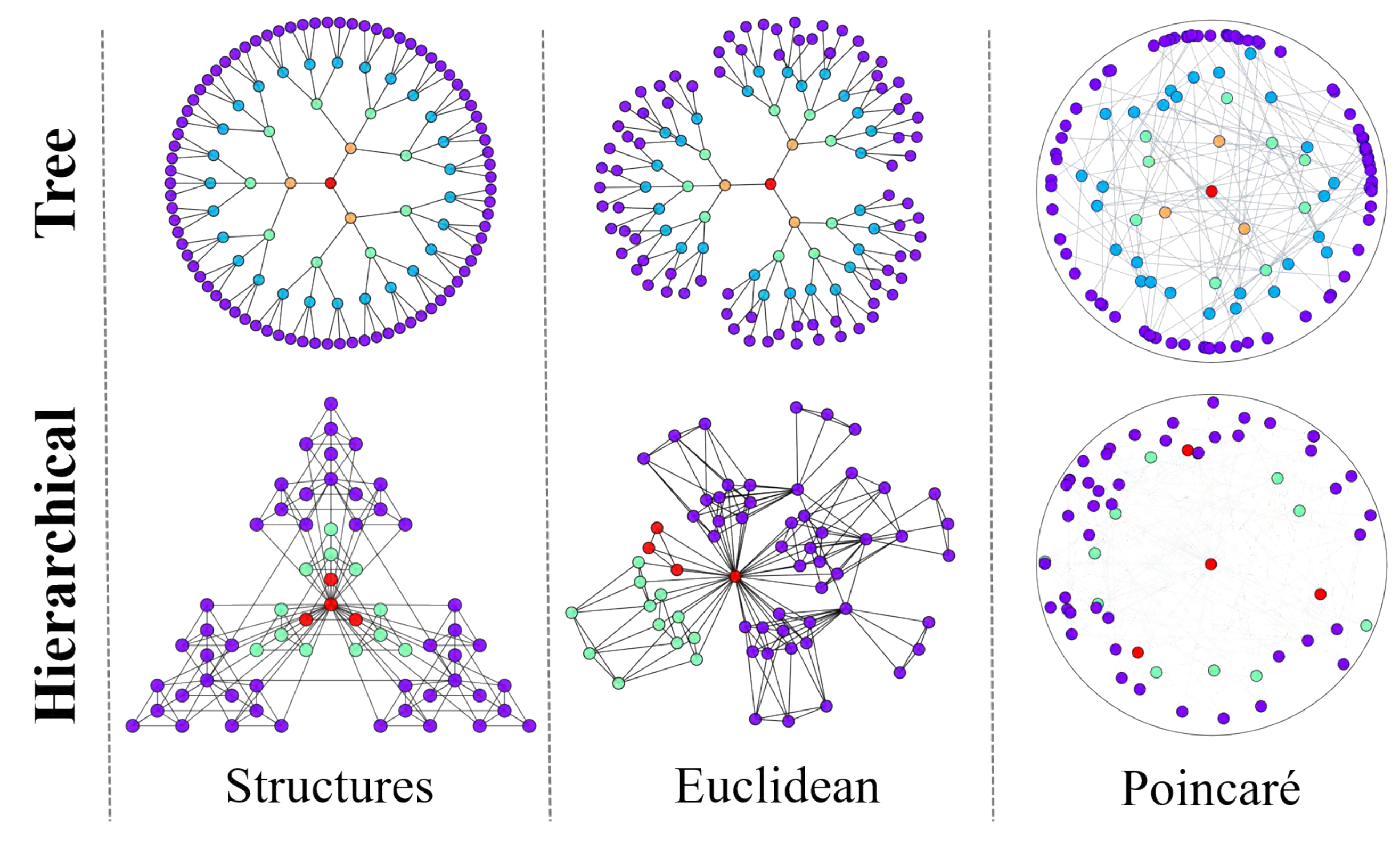}
%\caption{fig2}
}
\centering
\vspace{-1em}
\caption{(a) The quantitative analysis of the hierarchical graph and the Barab\'asi–Albert graph. 
With the growth of the network scale, the cluster coefficients of BA networks are referred to as the power-law distribution, and the cluster coefficients of the hierarchical network vary independently of the scale of the network. 
The connectivity correlation analysis shows that the nodes tend to connect nodes with similar connectivity and betweenness in BA networks, and the low-degree nodes of the hierarchical network tend to connect to the core nodes. 
(b) Compared with the Euclidean embedding method, the Poincar\'e Model can better represent the hierarchy of nodes in hyperbolic space. }
\vspace{-1em}
\label{fig:quantitative analysis}
\end{figure*}
\section{Preliminary}
In this section, we briefly introduce some notations and key definitions. 
Our work focuses on exploring the relationship between the imbalance issue and hierarchical geometric properties of labeled nodes in the semi-supervised node classification task on graphs. 

% Therefore, let us revisit the unbalanced semi-supervised node classification and the hyperbolic geometric model. 

\subsection{Semi-supervised Node Classification}
Given a graph $\mathcal{G}=\{\mathcal{V},\mathcal{E}\}$ with the node set  $\mathcal{V}$ of $N$ nodes and the edge set $\mathcal{E}$.  
Let $\mathbf{A}\in \mathbb{R}^{N\times N}$ be the adjacency matrix and $\mathbf{X}\in \mathbb{R}^{N\times d}$ be the node feature matrix, where $d$ denotes the dimension of node features. 
For node $v$, its neighbors set is $N (v): \{u \in \mathcal{V} |u, v \in \mathcal{E}\}. 
$ $d_v : |\mathcal{N} (v)|$ is the degree of node $v$.
Given the labeled node set $\mathcal{V}_L$ and their labels $\mathcal{Y}_L$ where each node $v_i$ is associated with a label $y_i$, \textit{semi-supervised node classification} aims to train a node classifier $f_{\theta}:v\rightarrow\mathbb{R}^{C}$ to predict the labels $\mathcal{Y}_U$ of remaining unlabeled nodes $\mathcal{V}_U=\mathcal{V} \setminus \mathcal{V}_L$, where $C$ denotes the number of classes. 
We separate the labeled node set $\mathcal{V}_L$ into $\{\mathcal{V}^{1}_L, \mathcal{V}^2_L, \cdots, \mathcal{V}^{C}_L\}$, where $\mathcal{V}^i_L$ denotes the nodes of class $i$ in $\mathcal{V}_L$. 
We focus on semi-supervised node classification based on GNNs methods.

%GNNs are mainly composed of message $\operatorname{M}$ and update function $\operatorname{U}$ in the message-passing paradigm, and the $l$th-GNNs layer is defined as: 

%\begin{equation}\label{equ:GNN_layer}
%   \begin{aligned}
%    \boldsymbol{h}_{i}^{l+1} &=\operatorname{U}^{l}\left(\boldsymbol{h}_{i}^{l}, \sum_{j \in \mathcal{N}(i)} \operatorname{M}^{l}\left(\boldsymbol{h}_{i}^{l}, \boldsymbol{h}_{j}^{l}, e_{i j}\right)\right),
%    \end{aligned}
%\end{equation}
%where $e_{i j}$ is the edge of nodes $\{v, u\} \in \mathcal{E}$.

\subsection{Hyperbolic Geometric Model}
Hyperbolic space is commonly referred to a manifold with constant negative curvature and is used for modeling complex networks. 
In hyperbolic geometry, five common isometric models are used to describe hyperbolic spaces~\cite{cannon1997hyperbolic}. 
% Among them, the Poincaré model has recently been introduced into machine learning for preserving the hierarchy of graphs. 
In this work, we use the Poincar{\'{e}} disk model to reveal the underlying hierarchy of the graph. 

\begin{myDef}[Poincar{\'{e}} disk Model]
	\label{def:poincare}
    The Poincar\'e Disk Model is a two-dimensional model of hyperbolic geometry with nodes located in the unit disk interior, and the generalization of n-dimensional with standard negative curvature $c$ is the Poincar\'e ball $\mathcal{B}^{n}_{c} = \{\boldsymbol{x} \in \mathbb{R}^n: \left\| \boldsymbol{x} \right\|^2 < 1/c \}$. For any point pair $(\boldsymbol{x}, \boldsymbol{y}) \in \mathcal{B}^{n}_c$, $\boldsymbol{x} \ne \boldsymbol{y}$, the distance on this manifold is defined as:
    \begin{equation}\label{Eq:hyperbolic_distance}
        d_{\mathcal{B}^{n}_{c}}(\boldsymbol{x}, \boldsymbol{y}) = \frac{2}{\sqrt{c}} \tanh^{-1} (\sqrt{c} \left\| -\boldsymbol{x} \oplus_c \boldsymbol{y} \right\|),
    \end{equation}
    where $\oplus_c$ is M\"{o}bius addition and $\left\| \cdot \right\|$ is $L_2$ norm.
\end{myDef}

\begin{myDef}[Poincar\'e norm]\label{def:Poincare_norm}
The Poincar{\'{e}} Norm is defined as the distance of any point $\boldsymbol{x} \in \mathcal{B}^{n}_{c}$ from the origin of Poincar{\'{e}} ball:
\begin{equation}
\label{Eq:distortion}
\left\| \boldsymbol{x} \right\|_{\mathcal{B}^{n}_{c}} = \frac{2}{\sqrt{c}} \tanh^{-1} (\sqrt{c} \left\| \boldsymbol{x} \right\|).
\end{equation}
\end{myDef}
% \begin{myDef}[Exponential and Logarithmic Maps]
%     The manifold $\mathcal{M}^d$ and the tangent space $\mathcal{T}_{\mathrm{x}}$ can be mapped to each other via \textit{exponential map} and \textit{logarithmic map}. 
%     The \textit{exponential map} $\mathrm{exp}_{\mathbf{x}}^{c}(\cdot)$ and \textit{logarithmic map} $\log _{\mathbf{x}}^{c}(\cdot)$ are defined as: 
%     \begin{equation}\label{log_exp_mapping}
%     \begin{aligned}
%         & \exp_{\boldsymbol{x}}^c(\boldsymbol{v}) = \boldsymbol{x} \oplus_c \frac{1}{\sqrt{c}} \tanh \left( \frac{\sqrt{c}\lambda_{\boldsymbol{x}}^c \left\| \boldsymbol{v} \right\|}{2} \right) \frac{\boldsymbol{v}}{\left\| \boldsymbol{v} \right\|}, \\
%         &\log_{\boldsymbol{x}}^c(\boldsymbol{u}) = \frac{2}{\sqrt{c}\lambda_{\boldsymbol{x}}^c} \tanh^{-1} \left( \sqrt{c} \left\| - \boldsymbol{x} \oplus_c \boldsymbol{u} \right\| \right) \frac{- \boldsymbol{x} \oplus_c \boldsymbol{u}}{\left\| - \boldsymbol{x} \oplus_c \boldsymbol{u} \right\|},
%     \end{aligned}
%     \end{equation}
%     where $\oplus_c$ is M\"{o}bius addition.
% \end{myDef}

%% file: 4_model.tex
\section{Understanding Hierarchy-imbalance}
In this section, we present a novel hierarchy-imbalance issue for semi-supervised node classification on graphs. 
Then a quantitative analysis of how the hierarchical nature of the graph affects the representation learning of nodes is presented. 
Finally, we present a new insight on the hierarchy-imbalance issue of graphs from a hyperbolic geometric perspective.

\subsection{Hierarchy-imbalance of Node Classification}
% GNNs learn node representations by aggregating valuable neighborhood information. 
The quantity and quality of the information received by a node determine the expressiveness of its representation in GNNs. 
In graphs with an intrinsic hierarchical structure, hierarchy is highly correlated with both the quantity and quality of information a node can receive. 
We argue that the imbalance of the node hierarchical tag affects the performance of GNNs in two aspects:

\noindent(1) \textbf{Implicit hierarchy-level: }
Considering the node label quality, the topological roles of labeled nodes are also highly relevant to the propagation of supervision information. 
Under the condition that the supervision information decays with the topological distance~\cite{buchnik2018bootstrapped}, the further the quality supervision information can achieve by propagation, the more significant influence the nodes can receive. 

\noindent(2) \textbf{Cross-hierarchy connectivity pattern: }
The hierarchical structure will introduce extra correlations in the graph topology, with the potential to cause nodes at different levels to have different patterns of neighborhood connectivity. 
The message-passing of supervision and other information may cause an over-squashing problem when the messages are across different hierarchy-levels with narrow connectivity~\cite{topping2021understanding}. 
The reason is that there are narrow "bottlenecks" between hierarchy-levels with different connectivity.

% Firstly, the hierarchical graph is a scale-free network with power-law degree distribution, and the number of nodes increases exponentially with the hierarchy increase. 
% Secondly, the hierarchy of a node represents the higher-level topological properties in the network, and nodes with higher hierarchy usually have better connectivity, more significant topological positions and play more important roles in a graph. 
% For example, the organization of an enterprise usually presents a hierarchical structure, and managers in departments who carry out decisions play a crucial role. 
% When we only have limited and imbalanced labeled information, the topological importance of nodes directly affects the quality of labeled information. 

\begin{figure*}[htb]
\centering
\includegraphics[width=0.90\textwidth]{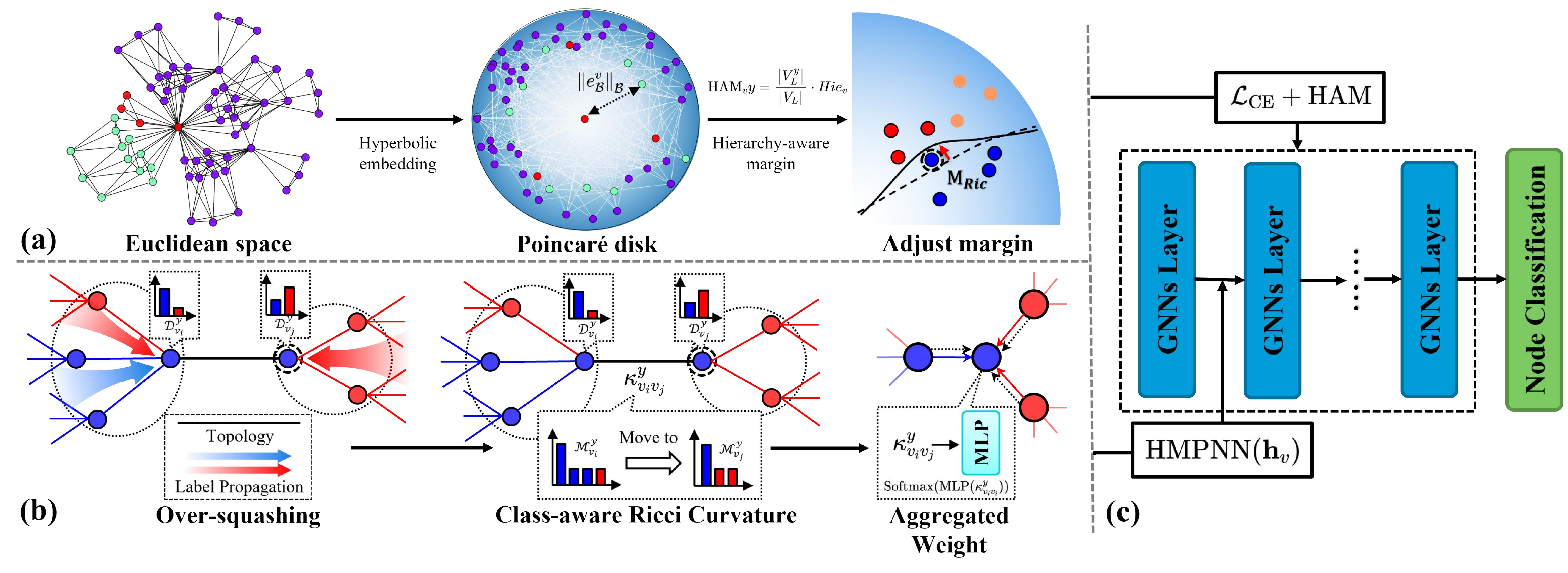}
\vspace{-1em}
\caption{\textbf{An illustration of \modelname~ architecture.} 
(1) \modelname~ learns the hyperbolic embedding and the hierarchy of each node by Poincar{\'{e}} model, then gets the hierarchy-aware margin and adds it to the GNNs loss to adjust the decision boundaries;
(2) \modelname~ calculates class-aware Ricci curvature for each edge, then transforms the Ricci curvatures to aggregated weights by an MLP and Softmax function. 
(3) \modelname~ performs as GNNs with HAM and HMPNN for the node classification. }
\vspace{-1em}
\label{fig:Architecture}
\end{figure*}

\subsection{Quantitative Analysis of Hierarchical Structure}
\label{Subsec:quantitative}
To further understand the hierarchy-imbalance issue, we use well-known graph models to generate two types of synthetic graphs for quantitative analysis: 
\textbf{(a) Hierarchical graph: } It is a deterministic fractal and scale-free graph with a 4-nodes module and 4-levels hierarchy and is generated by the Hierarchical Network Model~\cite{ravasz2003hierarchical}, 
\textbf{(b) Barab\'asi–Albert (BA) graph: } It is a random scale-free graph with power-law distributions and generated by the extended Barab\'asi–Albert Model~\cite{albert2000topology}. 
% We leverage these two synthetic graphs to simulate real-world complex networks with and without hierarchical structure, respectively. 

\noindent \textbf{Hierarchy and topological properties. }
Most classification methods based on graph topology rely on the modularity of graphs: the model can easily identify a set of nodes that are closely connected to each other within the class, but with few or no links to nodes outside the class~\cite{albert2000topology}.
% For example, the modules represent groups of friends or co-workers in a social network~\cite{albert2000topology}.
These clearly identifiable modular organizations have intuitive decision boundaries on the topology (Figure~\ref{fig:example} (b)). 
As shown in Figure~\ref{fig:quantitative analysis} (a) left plot, unlike the BA scale-free graph model in which the clustering coefficients are independent of the degree of a particular node, the clustering coefficients in a hierarchical network can be expressed by a function of degree, i.e., $C(k) = k^{-\beta}$, and the exponent $\beta = 1$ in deterministic scale-free networks~\cite{dorogovtsev2002pseudofractal}. 
It indicates that the decision boundaries may be implicit in the hierarchical topology (Figure~\ref{fig:example} (b)). 
% For the semi-supervised node classification task, the topological distance from the labeled nodes to the decision boundaries in the hierarchical graph is difficult to measure. 
To sum up, \textit{the topological role importance of labeled nodes in the hierarchy can more effectively affect the decision boundaries between node classes}.

\noindent\textbf{Hierarchy and correlations. }
The nodes with different hierarchy-levels have different connection patterns on the hierarchical graph. 
% , such as the nodes on the top-level of hierarchy making possible communication between local groups. 
To quantitatively analyze the local topological properties of nodes in the hierarchical network to reveal the connection patterns of different nodes, we consider two important local topological properties, connectivity(degree) and betweenness of nodes, where high connectivity represents nodes that are easier to propagate information, and high betweenness represents nodes that are on the "backbone" paths of the graph. 
To quantify the graph propagation patterns across levels of hierarchy, we compute the corresponding average nearest neighbor connectivity $\left\langle k_{n n}\right\rangle$ and betweenness $\left\langle b_{n n}\right\rangle$ of nodes with connectivity $k$ and betweenness $b$ as: 
\begin{equation}
\begin{aligned}
\label{Eq:correlations}
  \left\langle k_{n n}\right\rangle=\sum_{k^{\prime}} k^{\prime} \operatorname{prob} \left(k^{\prime} \mid k\right),
  \left\langle b_{n n}\right\rangle=\sum_{b^{\prime}} b^{\prime} \operatorname{prob}\left(b^{\prime} \mid b\right), 
\end{aligned}
\end{equation}
where $k^{\prime}$ and $b^{\prime}$ are connectivity and betweenness of other nodes, respectively. 
The results are shown in Figure~\ref{fig:quantitative analysis} (a) right plots.
For BA graph (blue lines), both $\left\langle k_{n n}\right\rangle$ and $\left\langle b_{n n}\right\rangle$ show that nodes tend to connect with other nodes whose connectivity and betweenness are similar to themselves. 
For the hierarchical graph (red lines), the $\left\langle k_{n n}\right\rangle$ results indicate that the nodes are more likely to connect to nodes at other different levels, and the $\left\langle b_{n n}\right\rangle$ results reveal that more nodes on "backbone" paths are more frequently connected with the nodes in the local group. 
For example, these hierarchical properties on the Internet are likely driven by several additional factors, such as economic market demand.
In conclusion, \textit{the quantitative analysis shows that the local connectivity and betweenness are closely related to the hierarchy of nodes, and the topological bottleneck of the graph may exist between different hierarchy-levels, which aggravates the over-squashing problem in the message-passing of supervision information}.

\subsection{Hierarchy of Hyperbolic Geometry Perspective}
Hyperbolic space can be understood as smooth versions of trees abstracting the hierarchical organization of complex networks~\cite{Krioukov2010Hyperbolic}. 
Figure~\ref{fig:quantitative analysis} (b) shows the node embeddings of the tree, hierarchical graphs on Euclidean and hyperbolic space. 
We can observe that the graph size grows as the radius of the Poincar\'e disk increases, and the hierarchy deepens as the graph size grows in hyperbolic space. 
Even though the hierarchical graph has a more complex structure, the position distribution of its hyperbolic embeddings is similar to a tree in hyperbolic space. 
% , which indicates that hyperbolic geometric embedding can effectively represent the underlying hierarchical structure of graph.
% In addition, for the hierarchical, tree-like structure of the underlying hyperbolic space, the graph's hierarchical structure manifests itself in the hierarchy of node degrees, and in the degree-dependent amount of space that nodes cover by their connections. 
In the GNNs community, learning the geometric properties of graphs has attracted much attention, and a typical case is learning the hierarchical structure of graphs using hyperbolic geometry~\cite{kennedy2016hyperbolicity,Octavian2018HyperbolicNeuralNetworks,NickelK17Poincare}. 
In summary, \textit{hyperbolic geometry provides us with exciting ways to capture and measure the implicit hierarchical structure of graphs. }

\section{\modelname~Model}
In this section, we present a novel hyperbolic geometric hierarchy-imbalance learning (\modelname) training framework to address the two main challenges of hierarchy-imbalance. 
The key insight is that we leverage hyperbolic geometry to abstract the implicit hierarchy of nodes in the graph and introduce a discrete geometric metric to deal with the over-squashing problem of supervision information propagated between hierarchy-levels. 
The architecture is shown in Figure~\ref{fig:Architecture}, and the overall process of \modelname~is shown in Algorithm~\ref{Alg:training}. 

\subsection{Hyperbolic Hierarchy-aware Margin}
% In the GNNs community, learning geometric properties of graphs has attracted much attention, and a typical case is learning the hierarchical structure of graphs using hyperbolic geometry~\cite{kennedy2016hyperbolicity,Octavian2018HyperbolicNeuralNetworks,NickelK17Poincare}. 
Our goal is to capture the implicit hierarchy of each labeled node, which is an important global property in a hierarchical graph to adjust the decision boundaries in the learning process. 
To this end, we design the Hyperbolic Hierarchy-Aware Margin (HAM), which consists of three steps:
First, we use the topological information of the graph to learn a hyperbolic embedding of the graph by using the Poincar\'e model. 
The hierarchical weights of nodes are then learned using their hyperbolic embeddings. Finally, a hyperbolic level-aware margin is designed to modify the objective function. 

\noindent \textbf{Step-1: Hyperbolic Embedding of Labeled Nodes. }
Poincar\'e embedding~\cite{NickelK17Poincare} is a shallow method of learning embedding into an $n$-dimensional Poincar\'e ball $\mathcal{B}^{n}_{c}$. 
In our work, we utilize Poincar\'e embedding to find the optimal embeddings of nodes by minimizing a hyperbolic distance-based loss function.
Based on the hyperbolic distance in Equation~\ref{Eq:hyperbolic_distance}, the loss function of Poincar\'e embedding is defined as follows: 
\begin{equation}\label{Eq:poincare_loss}
\begin{aligned}
    \mathcal{L}_{\mathcal{B}^{n}_{c}}(\Theta)=\sum_{(u, v) \in \mathcal{E}} \log \frac{e^{-d_{\mathcal{B}^{n}_{c}}(\boldsymbol{u}, \boldsymbol{v})}}{\sum_{\boldsymbol{v}^{\prime} \in \mathrm{Neg}(u)} e^{-d_{\mathcal{B}^{n}_{c}}\left(\boldsymbol{u}, \boldsymbol{v}^{\prime}\right)}},
\end{aligned} 
\end{equation}
where the negative examples $\mathrm{Neg}(u)$ of $u$ is $\mathrm{Neg}(u)=\{v |{u, v} \notin \mathcal{E}\} \cup\{u\}$. 
Then we utilize the stochastic Riemannian optimization method to solve the optimization problem as:
\begin{equation}\label{Eq:poincare_optimization}
    \Theta^{\prime} \leftarrow \underset{\Theta}{\arg \min } \mathcal{L}(\Theta) \quad \text { s.t. } \forall \theta_{i} \in \Theta :\left\|\theta_{i}\right\| < {1/c}.
\end{equation}
We follow Poincar\'e embedding using Riemannian stochastic gradient descent~\cite{bonnabel2013stochastic} to update the model parameters. 
For each labeled node $v \in \mathcal{V}_{L}$, we get the hyperbolic embedding $e^{v}_{\mathcal{B}}$ by Poincar\'e embedding method to capture the hierarchy of the node. 

\noindent \textbf{Step-2: Hyperbolic Hierarchy-aware Margin. }
In hyperbolic geometric space, the hyperbolic distance (radius) $R$ of an embedded node from the hyperbolic disk origin (North Pole) is able to abstract the depth of the hidden tree-like hierarchy~\cite{Krioukov2010Hyperbolic}. 
In our work, we compute the hyperbolic radius according to Equation~\ref{def:Poincare_norm} as the hierarchy of nodes by computing the Poincar\'e norm of the hyperbolic node embedding, and then we use a Multi-layer Perceptron (MLP) to transform the Poincar\'e norm into the hierarchy weights of the nodes. 
For each node $v \in \mathcal{V}$, the Hierarchy-aware Margin is defined as:
\begin{equation}\label{Eq:HAM}
   \operatorname{HAM}_{v^y} = \frac{|\mathcal{V}^{y}_{L}|}{|\mathcal{V}_L|} \operatorname{Softmax}\left(\operatorname{MLP} \left ( \left \|e^{v}_{\mathcal{B}} \right \|_{\mathcal{B}}\right)\right).
\end{equation}

\noindent \textbf{Step-3: Objective Function Adjustment. }
% Next, we introduce the efficient decision boundaries margin adjustment strategy HAM, which adaptively determines the intensity of the hierarchy-imbalance compensation based on the global hierarchy-level of an individual labeled node.
In Section~\ref{Subsec:quantitative}, we observe the hierarchy of a labeled node represents its global topological role and importance. 
Inspired by the margin-based imbalance handling methods~\cite{song2022tam}, we design the Hierarchy-Aware Margin to adaptively handle the intensity of supervision information based on a hierarchy to adjust the decision boundaries. 
The \modelname~learning objective function is formulated as:
\begin{equation}
\label{Eq:loss}
\begin{aligned}
    \mathcal{L} = \frac{1}{\|\mathcal{V}\|} \sum_{v \in \mathcal{V}} \mathcal{L}_{\operatorname{GNNs}} \left ( \operatorname{HMPNN}(\boldsymbol{h}_v) + \alpha \mathrm{HAM} , y_v \right ),
\end{aligned}
\end{equation}

\subsection{Hierarchy-aware Message-passing}
Although HAM can adjust the intensity of the supervision information on global topology, it cannot recognize the hierarchical connectivity patterns of individual nodes. 
Based on the observations in Section~\ref{Subsec:quantitative}, we draw a conclusion that a node of hierarchical graph tends to connect nodes with different connectivity (degree) and betweenness, and this cross-hierarchy connectivity pattern is more likely to lead to topology bottlenecks in message-passing. 

\noindent \textbf{Class-aware Ricci curvature.}
Recently, Ricci curvature has been introduced to analyze and measure the over-squashing problem caused by topological bottlenecks~\cite{topping2021understanding}. 
Inspired by this work, we extend \textit{Ollivier-Ricci curvature}~\cite{ollivier2009ricci} as the edge weights to affect message-passing, which can alleviate the over-squashing problem. 
Specifically, we first consider the label $i$ distribution in the one-hop neighborhood of a node $u$ is defined as:
\begin{equation}
\label{Eq:class-distribution}
\begin{aligned}
    D_{u, i}=\frac{\left|\left\{v \in \mathcal{N}(u) \mid y=i\right\}\right|}{\|\mathcal{N}(u)\|} .
\end{aligned}
\end{equation}
Our class-aware Ricci curvature $\kappa(u,v)_c$ of the edge $(u,v)$ is defined as:
\begin{equation}
\label{Eq:ricci-curvature}
\begin{aligned}
    \kappa(u,v) &= \frac{W(m^{y_u}_{u},m^{y_v}_{v})}{d(u,v)}, 
\end{aligned}
\end{equation}
where $W(\cdot,\cdot)$ is the Wasserstein distance, $d(\cdot,\cdot)$ is the geodesic distance (embedding distance), and $m^{y_u}_u$ is the mass distribution of node $u$. 
The mass distribution represents the important distribution of a node and its one-hop neighborhood~\cite{ollivier2009ricci}, and we further consider the label distribution in the neighborhood as:
\begin{equation*}
m^{\alpha,p}_u(u_i,D^{y_{u_i}}_{u_i})=
\begin{cases}
\alpha & \mbox{ if } u_i = x\\
\frac{1-\alpha}{C}\cdot  b^{-D^{y_{u_i}}_{u_i}d(u,u_i)^p} & \mbox{ if } u_i \in \mathcal{N}(u)\\
0 & \mbox{ otherwise }
\end{cases},
\end{equation*}
where $C=\sum_{u_i \in \mathcal{N}(u)} b^{-d(u,u_i)^p}$. 
$\alpha$ and $p$ are hyper-parameters that represent the importance of node $x$ and we take $\alpha = 0.5$ and $p = 2$ following the existing works~\cite{sia2019ollivier, ye2019curvature}.
 
\begin{algorithm}[t]
    \LinesNumbered
    \caption{\modelname} 
    \label{Alg:training}
    \KwIn{Graph $\mathcal{G}=\{\mathcal{V},\mathcal{E}\}$ with node labels $\mathcal{Y}$; Number of training epochs $E$; Loss hyperparameter $\alpha$}
    Parameter $\theta$ initialization;\\
    \KwOut{Predicted label $\hat{\mathcal{Y}}$.}
    \tcp{Learn hierarchy of nodes}
    Learning node Poincar\'e embedding $e_{\mathcal{B}} \gets$ Equation~\eqref{Eq:poincare_loss};\\
    Calculate node label distribution ${D} \gets$ Equation~\eqref{Eq:class-distribution};\\
    Calculate class-aware Ricci curvature $\kappa \gets$ Equation~\eqref{Eq:ricci-curvature};\\
    \For{$e=1,2,\cdots,E$}{
        \tcp{Learn hyperbolic hierarchy-imbalance margin}
        Calculate the $\operatorname{HAM} \gets$ Equation~\eqref{Eq:HAM};\\
        \tcp{Hierarchy-aware message-passing}
        Learning curvature-aware $\operatorname{HMPNN} \gets$ Eq.~\eqref{Eq:MPNN};\\
        Predict node labels $\hat{\mathcal{Y}}$ $\gets $ Eq.~\eqref{Eq:loss};\\
        \tcp{Optimize}
        Calculate the classification loss $\mathcal{L}\gets$ Eq.~\eqref{Eq:loss}, \\
        Update model parameters $\theta$ $\gets \theta-\eta \nabla \theta$. 
    }    
\end{algorithm}

\noindent \textbf{Curvature-Aware Message-Passing.}
Class-aware Ricci curvature measures how easily the label information flows through an edge and can be used to guide message-passing. 
We follow~\cite{ye2019curvature} by using an MLP to learn the mapping function from the curvature to the aggregated weights $\tau_{u v}$ of MPNN.
We have Hierarchy-aware Massage-Passing Neural Networks (HMPNN) as follow:
\begin{equation}
\begin{aligned}
    \label{Eq:MPNN}
    \tau_{u v} &= \operatorname{Softmax}\left(\operatorname{MLP} \left ( \kappa(u,v) \right)\right),\\
    \boldsymbol{ h}_{i}^{l+1} &=U^{l}\left(\boldsymbol{h}_{i}^{l}, \sum_{j \in \mathcal{N}(i)} M^{l}\left(\boldsymbol{h}_{i}^{l}, \boldsymbol{h}_{j}^{l}, \tau_{i j} e_{i j}\right)\right). 
\end{aligned}
\end{equation}

% We leverage curvature parameter $\tau_{u v}$ re-weight neighborhood edges of node $u$, and directly affecting the message-passing of label information. 
% We take edge weights into Equation~\ref{equ:GNN_layer} and have Hierarchy-aware Massage-Passing Neural Networks (HMPNN) as follow:

% \subsection{Objective Function and Optimization}
% The overall architecture is shown in Figure~\ref{fig:Architecture}. 
% where $\mathcal{L}$ is the loss function (such as cross entropy), and $\alpha$ is hyperparameter. 
% Note that our method simply re-weights the message passing and adjusts the decision boundaries on the final loss function by using the implicit geometric properties of the graph topology, so our method can be combined orthogonally with any GNNs method.
% The overall process of \modelname~is shown in Algorithm~\ref{Alg:training}. 

%% file: 5_experiment.tex
\begin{table}[t]
\small
\caption{Statistics of real-world datasets.}
\vspace{-1em}
\centering

\begin{tabular}{cl|ccccc}
\toprule
\multicolumn{2}{c|}{\textbf{Dataset}}  & \textbf{\#Node} & \textbf{\#Edge} & \textbf{\#Label} &\textbf{\#Avg. Deg} & \textbf{\#$\mathcal{H}$($\mathcal{G}$)} \\ 
\midrule
&\textbf{Cora}      & 2,708 & 5,429 & 7 & 4.01   &0.83     \\
&\textbf{Citeseer}  & 3,327 & 4,732 & 6  & 2.85  &0.72       \\
&\textbf{Photo}     & 7,487 & 119,043 & 8  & 31.80  &0.83   \\
&\textbf{Actor}     & 7,600  & 33,544 & 5   & 8.83  &0.24  \\
&\textbf{Chameleon} & 2,277  & 31,421 & 5  & 27.60  &0.25   \\
&\textbf{Squirrel}  & 5,201  & 198,493 & 5  & 76.33  &0.22   \\
\bottomrule
\end{tabular}

\label{dataset_description}
\end{table}

\section{Experiment}
\label{sec:Experiment}
In this section, we conduct comprehensive experiments to demonstrate the effectiveness and adaptability of \modelname~\footnote{The code is available at \url{https://github.com/RingBDStack/HyperIMBA}.} on various datasets and tasks. 
We further analyze the robustness to investigate the expressiveness of \modelname. 

\begin{table*}[htbp]
\caption{Weighted-F1 score and Micro-F1 score (\% ± standard deviation) of node classification on real-world graph datasets.\\
(Result: average score ± standard deviation; \textbf{Bold}: best; \underline{Underline}: runner-up; The hyphen symbol indicates experiments results are not accessible due to memory issue or time limits.)}
\vspace{-1em}
\centering
\resizebox{\textwidth}{!}{
\begin{tabular}{cccccccccccccc}
\hline
 &
  \multirow{2}{*}{Model} &
  \multicolumn{2}{c}{Cora} &
  \multicolumn{2}{c}{Citeseer} &
  \multicolumn{2}{c}{Photo} &
  \multicolumn{2}{c}{Actor} &
  \multicolumn{2}{c}{Chameleon} &
  \multicolumn{2}{c}{Squirrel} \\ \cline{3-14}
  &
   &  W-F1 &  M-F1 &  W-F1 &  M-F1 &  W-F1 &  M-F1 &  W-F1 &  M-F1 &  W-F1 &  M-F1 &  W-F1 &  M-F1 
  \\ \hline
\multirow{6}{*}{\rotatebox{90}{GCN}} &
  original &79.4±0.9   &77.5±1.5   &66.3±1.3   &62.2±1.2   &85.4±2.8   &84.6±1.3   &21.8±1.3   &20.9±1.4   &30.5±3.4   &30.5±3.3   &21.9±1.2   &21.9±1.2   \\
 &
  ReNode &80.0±0.7   &78.4±1.3   &66.4±1.0   &62.4±1.1   &86.2±2.4   &85.3±1.6   &21.2±1.2   &20.2±1.6   &30.3±3.2   &30.4±2.8   &22.4±1.1   &22.4±1.1   \\
 &
  DropEdge &79.8±0.8   &77.8±1.0   &66.6±1.4   &63.4±1.6   &86.8±1.7   &85.4±1.3   &22.4±1.0   &21.4±1.3   &30.6±3.5   &30.6±3.3   &\underline{22.8±1.2}   &\underline{22.8±1.2}   \\
 &
  SDRF &\underline{82.1±0.8}   &\underline{79.3±1.0}   &\underline{69.6±0.4}   &\underline{66.6±0.3}   &-   &-   &-   &-   &\underline{32.3±0.7}   &\underline{39.0±1.2}   &-   &-   \\
   &
  ReNode+TAM &80.1±0.9 &78.2±1.6   &67.1±1.4  &62.3±0.9   &\underline{87.6±1.3}   &\underline{86.9±1.0}   &\underline{23.1±0.9}   &\underline{22.2±1.3}   &\underline{32.3±0.9}  &32.1±0.8   &22.1±0.4    & 22.1±0.3   \\
   &
  \textbf{\modelname}  &\textbf{83.0±0.3}  &\textbf{83.1±0.4}   &\textbf{76.3±0.2}   &\textbf{73.4±0.3}   &\textbf{92.8±0.3}   &\textbf{92.5±0.3}   &\textbf{30.7±0.2}  &\textbf{29.3±0.4}   &\textbf{44.1±0.7}   &\textbf{42.3±1.1}   &\textbf{31.2±2.4}   &\textbf{28.4±2.0}  \\
  \hline
\multirow{6}{*}{\rotatebox{90}{GAT}} &
  original  &78.3±1.5   &76.4±1.7   &64.4±1.7   &60.6±1.7   &88.2±2.9   &86.2±2.6   &21.8±1.2   &20.9±1.1   &29.9±3.5   &29.9±3.1   &20.5±1.4   &20.5±1.4   \\
 &
  ReNode  &\underline{78.9±1.2}   &77.2±1.5   &\underline{64.9±1.6}   &61.0±1.5   &\underline{89.1±2.4}   &87.1±2.6   &21.5±1.2   &20.5±1.1   &29.2±2.3   &29.1±2.0   &20.4±1.8   &20.4±1.8   \\
 &
  DropEdge  &78.7±1.3   &76.9±1.5   &64.5±1.4   &60.5±1.3   &88.9±1.9   &87.1±2.1   &\underline{22.9±1.2}   &\underline{21.8±1.1}   &30.3±1.6   &30.2±1.2   &\underline{21.2±1.5}   &\underline{21.2±1.5}   \\
 &
  SDRF  &77.9±0.7   &75.9±0.9   &\underline{64.9±0.6}   &61.9±0.9   &-   &-   &-   &-   &\underline{43.0±1.9}   &\underline{42.5±1.9}   &-   &-   \\
 &
  ReNode+TAM &78.4±1.3 &\underline{77.3±1.3}   &64.2±1.3   &\underline{63.1±0.8}   &89.0±1.8   &\underline{87.3±1.7}   &21.3±1.2    &20.7±1.1   &30.9±1.5   &30.2±1.8   &20.0±1.4 &19.4 ±1.2  \\
 &
  \textbf{\modelname}  &\textbf{83.5±0.3}  &\textbf{83.6±0.3}   &\textbf{75.0±0.4}   &\textbf{73.0±0.4}   &\textbf{92.5±0.5}   &\textbf{92.1±0.8}   &\textbf{30.9±1.0}   &\textbf{29.8±1.0} &\textbf{43.2±0.7}   &\textbf{42.5±0.6}   &\textbf{31.1±1.0}   &\textbf{28.7±1.3}  \\
  \hline
\multirow{6}{*}{\rotatebox{90}{GraphSAGE}} &
  original  &75.4±1.6  &74.1±1.6   &64.8±1.6   &60.7±1.6   &86.1±2.5   &83.3±2.4   &24.0±1.2   &23.2±1.0   &36.5±1.6   &36.2±1.6   &27.2±1.7   &27.2±1.7  \\
 &
  ReNode  &\textbf{76.4±0.9}  &\textbf{75.0±1.1}   &65.4±1.7   &61.2±1.7   &\textbf{86.5±1.7}   &\textbf{84.1±1.7}   &23.7±1.2   &22.8±1.0   &36.4±1.9   &36.1±1.9   &27.7±1.8   &27.7±1.8  \\
 &
  DropEdge  &\underline{76.0±1.6}  &74.5±1.6   &65.1±1.4   &60.9±1.4   &86.2±1.6   &83.5±1.4   &\underline{24.1±1.0}   &\underline{23.3±0.9}   &37.5±1.4   &37.2±1.4   &27.5±1.8   &27.5±1.8  \\
 &
  SDRF  &75.7±0.8  &74.6±0.8   &65.3±0.6   &61.4±0.6   &-   &-   &-   &-   &\underline{41.5±2.6}   &\underline{41.6±2.7}   &-   &-  \\
 &
  ReNode+TAM &\underline{76.0±1.1} &\underline{74.9±1.0}   &\underline{67.1±2.0}  &\underline{63.4±1.2}   &\underline{86.4±1.4}   &\underline{83.8±1.2}   &23.6±1.2   &22.5±1.3   &38.3±1.8  &38.1±1.8   &\underline{27.8±1.4} & \underline{27.8±1.4}   \\
 &
  \textbf{\modelname}  &72.4±0.3  &71.5±0.5   &\textbf{72.8±0.2}   &\textbf{70.5±0.3}   &80.9±1.2   &78.2±1.2   &\textbf{35.6±0.6} &\textbf{34.3±1.1}   &\textbf{42.9±0.6}   &\textbf{42.5±0.6}   &\textbf{38.6±1.1}   &\textbf{36.9±0.7}   \\ \hline
  \end{tabular}
}
\vspace{-0.5em}
\label{table:results}
\end{table*}

\begin{table*}
\caption{Weighted-F1 scores (\% ± standard deviation) and improvements (\%) results of Ablation Study. (Result: average score ± standard deviation; \textbf{Bold}: best.)}
\vspace{-1em}
\label{tab:LMH}
\resizebox{\linewidth}{!}{%
\centering
\begin{tabular}{ccccccccccccc}
\hline
\multirow{2}{*}{Model}
& \multicolumn{2}{c}{Cora} 
& \multicolumn{2}{c}{Citeseer} 
& \multicolumn{2}{c}{Photo}
& \multicolumn{2}{c}{Actor}
& \multicolumn{2}{c}{Chameleon}
& \multicolumn{2}{c}{Squirrel}
\\ \cline{2-13}

& W-F1 (\%)  & $\Delta$ (\%)  
& W-F1 (\%)  & $\Delta$ (\%)  
& W-F1 (\%)  & $\Delta$ (\%)  
& W-F1 (\%)  & $\Delta$ (\%)  
& W-F1 (\%)  & $\Delta$ (\%)  
& W-F1 (\%)  & $\Delta$ (\%)  
\\ \cline{1-13}
GCN                     &79.4±0.9   &-              &66.3±1.3   &-              &85.4±2.8                   &-                  &21.8±1.3   &-              &30.5±3.4   &-              &21.9±1.2   &-  \\
\modelname~(w/o HMPNN)    &82.3±0.3   &$\uparrow$2.9  &71.3±0.7   &$\uparrow$5.0  &92.4±0.3   &$\uparrow$7.0   &29.6±2.9   &$\uparrow$7.8  &39.6±0.9   &$\uparrow$9.1  &24.9±0.8   &$\uparrow$3.0  \\
\modelname~(w/o HAM) &82.9±0.5   &$\uparrow$3.5  &75.8±0.5   &$\uparrow$9.5  &92.6±0.4       &$\uparrow$7.2      &30.1±0.5   &$\uparrow$8.3  &42.4±0.9   &$\uparrow$11.9 &26.3±0.8   &$\uparrow$4.4  \\
\textbf{\modelname~}        &\textbf{83.0±0.3}  &$\uparrow$\textbf{3.6} &\textbf{76.3±0.2}   &$\uparrow$\textbf{10.0}    &\textbf{92.8±0.3}  &$\uparrow$\textbf{7.4} &\textbf{30.7±0.2}   &$\uparrow$\textbf{8.9}     &\textbf{44.1±0.7}  &$\uparrow$\textbf{13.6}&\textbf{31.2±2.4}   &$\uparrow$\textbf{9.3}  \\
\hline
\end{tabular}
}
\vspace{-0.5em}
\label{tab:ablation}
\end{table*}

\subsection{Datasets}
We conduct experiments on synthetic and real-world datasets to evaluate our method, and analyze the model's capabilities in terms of both graph theory and real-world scenarios. 
The statistics of the datasets are summarized in Table~\ref{dataset_description}. The edge homophily $\mathcal{H}(\mathcal{G}) $ is computed according to \cite{peng2020GemoGCN_HG}.

%The computation of edge homophily\cite{peng2020GemoGCN_HG} across neighborhoods and normalization as follow: 
%\begin{equation}
%\label{Eq:homophily}
%\begin{aligned}
%    \mathcal{H}(\mathcal{G}) = \frac{1}{|\mathcal{V}|} \sum_{v\in\mathcal{V} }\frac{\left | \left \{ (u,v):u\in N(v)\wedge y_{v} = y_{u} \right \}  \right | }{\left | N(v)  \right | } .
%\end{aligned}
%\end{equation}

\textbf{Synthetic Datasets.}
We generate the hierarchical synthetic graphs for essential verification and analysis of our method by the well-accepted graph theoretical model: \textbf{Hierarchical Network Model} (\textbf{HNM})~\cite{ravasz2003hierarchical}. 
For each dataset, we create 1,024 nodes and subsequently perform the graph generation algorithm on these nodes. 
For the hierarchical graph, we consider an initial network of $N=4$ fully interconnected nodes as the fractal, and derive $k=5$ times in an iterative way by replicating the initial fractal of the graph according to the fractal structure. 
For each generated graph, we randomly select 80\% nodes as the test set, 10\% nodes as the training set, and the other 10\% nodes as the validation set. 

\textbf{Real-world Datasets.}
% To further verify the advantages of \modelname, 
We also conducted experiments on several real-world datasets: 
(1) \textit{Citation network:} Cora and Citeseer~\cite{sen2008cora} are citation networks of academic papers. 
(2) \textit{Co-occurrence network:} Photo~\cite{shchur2018pitfalls} is segment of the Amazon co-purchase graph and Actor~\cite{pei2020geom} is an actor co-occurrence network.
(3) \textit{Page-page network:} Chameleon and Squirrel~\cite{rozemberczki2021multi} are page-page networks on Wikipedia. 
Since we focus on the imbalance issue of topological properties, we set the same number of labeled nodes for each class. 

\begin{figure}[!t]
\centering
\subfigure{
\includegraphics[width=0.98\columnwidth]{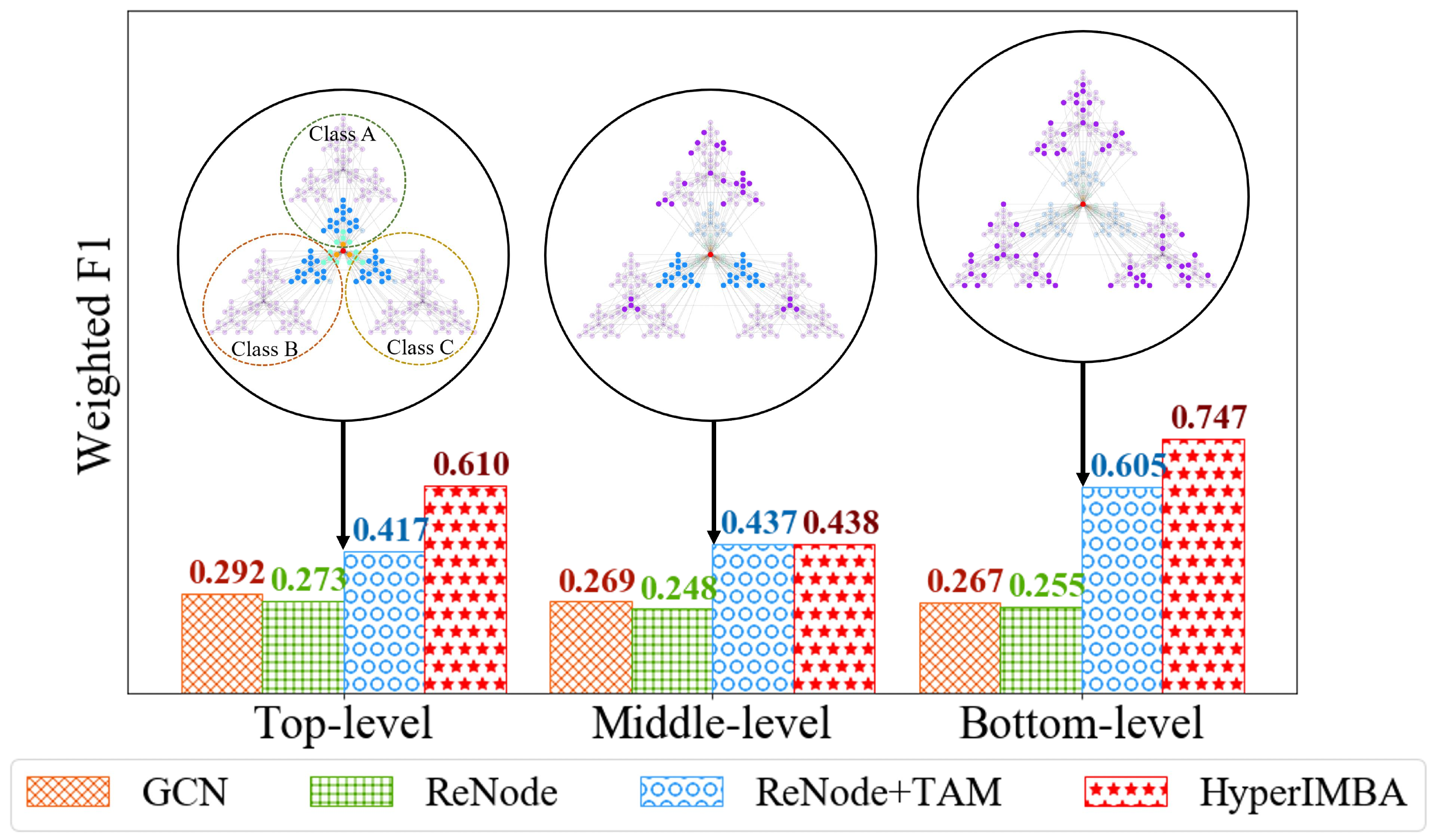} 
}
\vspace{-2em}
\caption{Performances of different hierarchy-level training setup on the synthetic graph. }
\vspace{-1em}
\label{fig:analysis_synth}
\end{figure}

\subsection{Experimental Setup}
\textbf{Baselines.} 
We choose well-known GNNs as backbones, including GCN~\cite{GCN}, GAT~\cite{GAT}, and GraphSAGE~\cite{hamilton2017inductive}. 
To evaluate the proposed \modelname, we compare it with a variety of baselines, including:
the most relevant baselines of the topology-imbalance issue are ReNode~\cite{chen2021topology} and TAM~\cite{song2022tam}. 
ReNode is a position-aware and re-weighted~\cite{cao2019learning,cui2019class,ren2018learning} method, and TAM is a neighborhood class information-aware margin-based method. 
DropEdge~\cite{rong2019dropedge} randomly removes a certain number of edges from the input graph at each training epoch, which acts as data augmentation~\cite{park2021graphens} in terms of structure. 
SDRF~\cite{topping2021understanding} is a structure rewiring method for the over-squashing issue, which modifies edges with Ricci curvatures. 

\textbf{Settings.} 
We set the depth of GNN backbones as 2 layers and adopt the implementations from the PyTorch Geometric Library\footnote{\url{https://github.com/rusty1s/pytorch_geometric}} in all experiments. 
We set the representation dimension of all baselines and \modelname~to 256. 
The parameters of baselines are set as the suggested value in their papers or carefully tuned. 
For DropEdge, we set the edge dropping/adding probability to 10\%. 
For \modelname, we set the hyperbolic curvature $c=1$ of Poinca\'e model. 

\subsection{Performance Evaluation}\label{sec:performance}
\textbf{Performance on Synthetic Graphs.}
To verify the hierarchy capturing ability, we evaluate our method on hierarchical organization synthetic graph HNM. 
The node classes of HNM are three communities with the same hierarchical structure and are evenly distributed in three directions of the graph, as shown in Figure~\ref{fig:analysis_synth}. 
We divide the hierarchical graph into three levels according to the hierarchy-level: the top-level (1, 2, 3-order fractals of HNM), the middle-level (4-order fractals of HNM), and the bottom-level (5-order fractals of HNM), respectively, to verify the effect of the model using labeled nodes at different hierarchy-levels as training samples. 
We allow randomly sample labeled nodes in low-level to supplement the high-level labeled nodes together as training nodes to reach a consensus for each class. 
According to Figure~\ref{fig:analysis_synth}, \modelname~ significantly outperforms all baselines, especially with top-level or bottom-level training setup. 
% It demonstrates the superiority of our \modelname~ in capturing the hierarchy of graphs. 
In addition, unlike the performance of ReNode and TAM increases monotonically from the top- to bottom-level, the performance of \modelname~and vanilla GCN is higher in the top- and bottom-level than in the middle-level. 
This phenomenon matches perfectly with the hierarchical connectivity pattern which we discussed in Section ~\ref{Subsec:quantitative}, i.e., nodes of a hierarchical graph tend to connect with nodes of different connectivity and betweenness rather than the nodes with similar properties. 
\modelname~benefits from the discrete curvature-aware re-weighting, which effectively alleviates the over-squashing problem caused by cross-hierarchy connectivity pattern. 

\begin{figure*}[!t]
\centering
\includegraphics[width=0.98\textwidth]{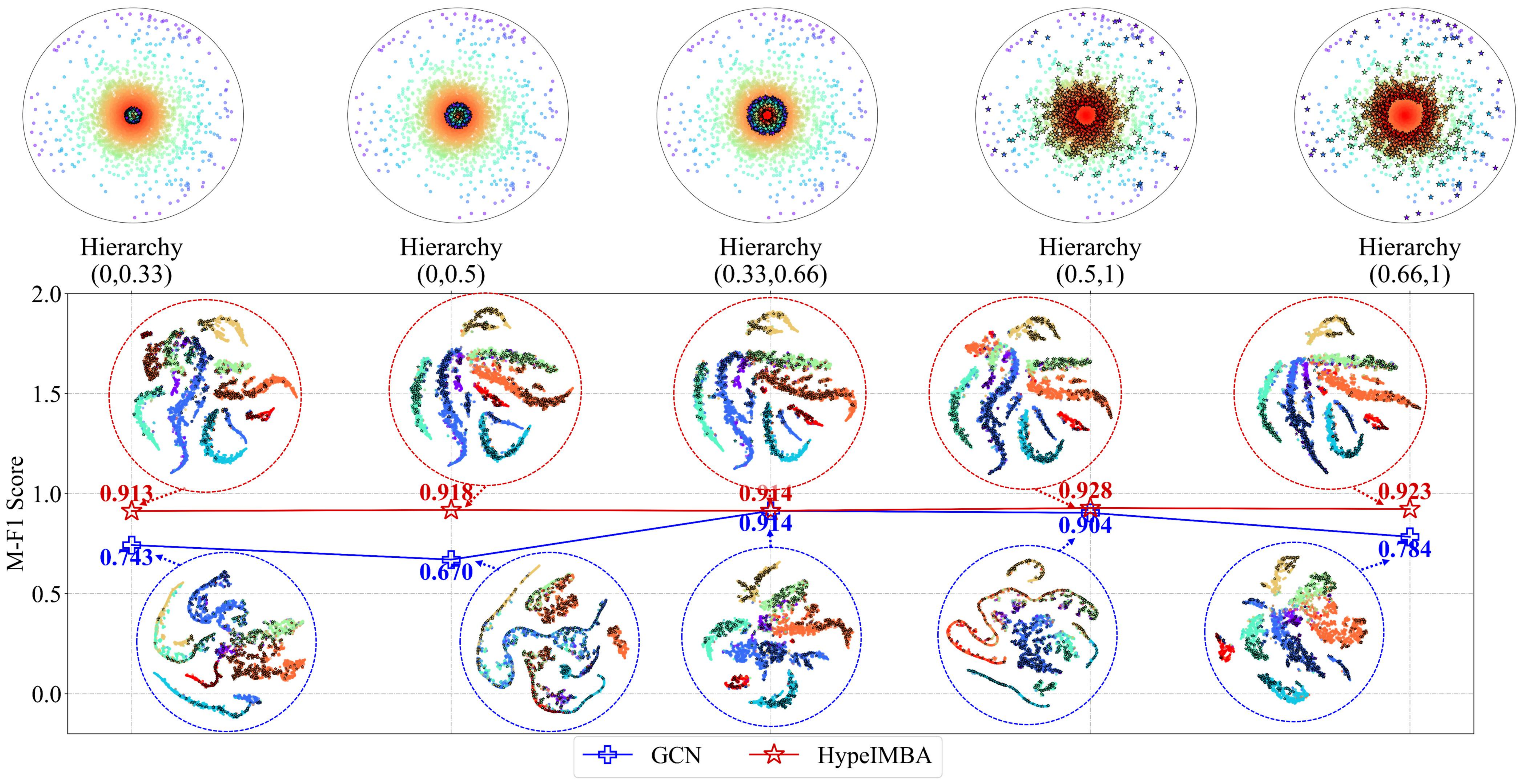} 
\vspace{-1em}
\caption{Performances and analyis on Photo with different hierarchy-levels training setting.}
\vspace{-1em}
\label{fig:analysis_real}
\end{figure*}
% \vspace{-1em}

\noindent\textbf{Performance on Real-world Graphs.}
Table~\ref{table:results} summarizes the performance of \modelname~ and all baselines on six real-world datasets. 
Our \modelname~shows significant superiority in improving the performance of GCN and GAT on all datasets.
It demonstrates that \modelname~ is capable of capturing the underlying topology and important connectivity patterns, especially for the backbones that can thoroughly learn the topology. 
Our method has only a few improvements for the backbone on high homophily and weak hierarchical graphs such as Cora. 
% The reason may be that our method focuses on exploiting the geometric priors and properties of the topology to improve the learning performance, which does not perform well . 
By contrast, our method achieves an overwhelming advantage on graphs with high heterophily datasets (Actor, Chameleon, and Squirrel). 
TAM improves the performance of ReNode by considering the label connectivity of the neighborhood, but it still does not work well on the graph with poor connectivity (Citeseer). 
The reason for the poor performance of ReNode is that topological boundaries are difficult to be directly used as decision boundaries on real-world graphs. 
%We think it’s because the global and higher-order topological properties may be able to be more useful than the neighborhood connectivity on heterophilic graphs. 
Compared with SDRF, \modelname~ further considers the over-squashing problem of supervision information, and the improvement of results also confirms our intuition. 
Note that the performance of \modelname~ depends on whether global and higher-order topological properties play an important role in learning. 
For the subgraph-sampling method (GraphSAGE), \modelname~ can still obtain significant improvement in most cases of incomplete topology information. 
% In addition, it can observe that our method does not work well on homophilous graphs by using subgraph sampling method as the backbone. 
% For the subgraph-sampling method (GraphSAGE), we think the specific sampling strategy may loses the high-order topological information uncontrollably, which leads to a huge distortion of the topological geometric prior introduced by \modelname. 

\subsection{Analysis of \modelname}
In this subsection, we conduct ablation studies for HAM and HMPNN, to provide further performance analysis for our model. 
Then we perform a case study of hierarchy-imbalance learning, and provide  the observation of the learning performances under different hierarchy-level training settings to explore the intrinsic mechanism of the hierarchy-imbalance issue. 
We also visualize the learning results to provide further insights into the impact caused by the hierarchy-imbalance issue more intuitively.

\noindent\textbf{Ablation Study.}
We conduct ablation studies for the two main mechanisms of \modelname, hierarchy-aware margin and hierarchy-aware message-passing.
We choose GCN as the backbone, and the results are shown in Table~\ref{tab:ablation}. HAM plays a key role in alleviating the hierarchy-imbalance issue, and demonstrate the superiority of hyperbolic geometry for capturing the underlying hierarchy of graphs. 
Moreover, HMPNN also significantly alleviates the over-compression problem of label supervision information. 
In summary, \modelname~consistently outperforms the GCN and the other two variants on all real-world datasets. 

\noindent\textbf{Case study and Visualization.}
We construct a case study based on the real-world dataset \textit{Photo} to explore how the labeled nodes at different hierarchy-levels will affect the learning models. 
% We first re-sample the training samples based on the default data split according to the hyperbolic embedding of \textit{Photo}. 
We divide five regions in the embedded Poincar\'e disk of \textit{Photo}, and randomly sample labeled nodes in the regions as training samples, respectively. 
In order to satisfy the quantity-balanced setting of labeled nodes for each node class, we also perform a random supplement selection of nodes according to a certain probability as in Subsection~\ref{sec:performance}. 
Figure~\ref{fig:analysis_real} shows the training setting of five hierarchy-levels and reports the performances and visualizations of GCN and \modelname~for each hierarchy-level using t-SNE\cite{van2008tsne}. 

As we can observe in Figure~\ref{fig:analysis_real}, the labeled nodes with different hierarchy-levels significantly affect the shapes and boundaries of the node embedding clusters, which indicates that the hierarchical properties can directly affect the decision boundary of the model by handling the connectivity pattern on the graph. 
An interesting observation is that the top-level labeled nodes make the embedding distribution much more compact and produce a large number of false positives, which indicates that it has a severe over-squashing problem in the message-passing process. 
It is consistent with the quantitative analysis in Section~\ref{Subsec:quantitative}, i.e., the nodes with high connectivity and betweenness refer to connect nodes with low connectivity and betweenness according to hierarchical connectivity patterns. 
In addition, we observe that the bottom-level nodes with more diverse information, resulting in node clusters with diffuse shapes and wider boundaries, may easily lead to conflicts or overlaps between different node classes. 
The visualization of the results in Figure~\ref{fig:analysis_real} shows that \modelname~consistently maintains the appropriate node cluster shapes and boundaries under different hierarchy-level training settings. 
% It indicates that \modelname~can adaptively achieve a perfect trade-off between over-squashing and decision boundary problems under different training settings. 
% In a nutshell, benefiting from the HAM and HMPNN mechanisms, the \modelname~again demonstrates its effectiveness. 

%% file: 6_conclusion.tex
\section{Conclusion}
In this paper, for the first time, we explore the hierarchy-imbalance issue on the hierarchical structure, which is a significant topological property of the graph. 
We proposed \modelname, a novel training framework to alleviate the hierarchy-imbalance issue from the hyperbolic geometric perspective. 
\modelname~can effectively capture the implicit hierarchy of nodes by hyperbolic geometric embedding, and we propose a hierarchy-aware margin for adjusting classification decision boundaries. % on the hierarchical structure of the graph. 
Moreover, \modelname~leverages discrete local curvature to improve the message-passing mechanism for alleviating the over-squashing problem caused by hierarchy connectivity patterns. 
Experimental results on synthetic and real-world datasets demonstrate that \modelname~consistently and significantly outperforms existing works.

%% file: 0_main.bbl

\begin{thebibliography}{51}

%%% the Web address.

\ifx \showCODEN    \undefined \def \showCODEN     #1{\unskip}     \fi
\ifx \showDOI      \undefined \def \showDOI       #1{#1}\fi
\ifx \showISBNx    \undefined \def \showISBNx     #1{\unskip}     \fi
\ifx \showISBNxiii \undefined \def \showISBNxiii  #1{\unskip}     \fi
\ifx \showISSN     \undefined \def \showISSN      #1{\unskip}     \fi
\ifx \showLCCN     \undefined \def \showLCCN      #1{\unskip}     \fi
\ifx \shownote     \undefined \def \shownote      #1{#1}          \fi
\ifx \showarticletitle \undefined \def \showarticletitle #1{#1}   \fi
\ifx \showURL      \undefined \def \showURL       {\relax}        \fi
% invisible to TeX
\providecommand\bibfield[2]{#2}
\providecommand\bibinfo[2]{#2}
\providecommand\natexlab[1]{#1}
\providecommand\showeprint[2][]{arXiv:#2}

\bibitem[\protect\citeauthoryear{Albert and Barab{\'a}si}{Albert and
  Barab{\'a}si}{2000}]%
        {albert2000topology}
\bibfield{author}{\bibinfo{person}{R{\'e}ka Albert} {and}
  \bibinfo{person}{Albert-L{\'a}szl{\'o} Barab{\'a}si}.}
  \bibinfo{year}{2000}\natexlab{}.
\newblock \showarticletitle{Topology of evolving networks: local events and
  universality}.
\newblock \bibinfo{journal}{\emph{Physical review letters}}
  \bibinfo{volume}{85}, \bibinfo{number}{24} (\bibinfo{year}{2000}),
  \bibinfo{pages}{5234}.
\newblock


\bibitem[\protect\citeauthoryear{Albert, Jeong, and Barab{\'a}si}{Albert
  et~al\mbox{.}}{1999}]%
        {albert1999diameter}
\bibfield{author}{\bibinfo{person}{R{\'e}ka Albert}, \bibinfo{person}{Hawoong
  Jeong}, {and} \bibinfo{person}{Albert-L{\'a}szl{\'o} Barab{\'a}si}.}
  \bibinfo{year}{1999}\natexlab{}.
\newblock \showarticletitle{Diameter of the world-wide web}.
\newblock \bibinfo{journal}{\emph{nature}} \bibinfo{volume}{401},
  \bibinfo{number}{6749} (\bibinfo{year}{1999}), \bibinfo{pages}{130--131}.
\newblock


\bibitem[\protect\citeauthoryear{Barab{\'a}si and Albert}{Barab{\'a}si and
  Albert}{1999}]%
        {barabasi1999emergence}
\bibfield{author}{\bibinfo{person}{Albert-L{\'a}szl{\'o} Barab{\'a}si} {and}
  \bibinfo{person}{R{\'e}ka Albert}.} \bibinfo{year}{1999}\natexlab{}.
\newblock \showarticletitle{Emergence of scaling in random networks}.
\newblock \bibinfo{journal}{\emph{Science}} \bibinfo{volume}{286},
  \bibinfo{number}{5439} (\bibinfo{year}{1999}), \bibinfo{pages}{509--512}.
\newblock


\bibitem[\protect\citeauthoryear{Bonnabel}{Bonnabel}{2013}]%
        {bonnabel2013stochastic}
\bibfield{author}{\bibinfo{person}{Silvere Bonnabel}.}
  \bibinfo{year}{2013}\natexlab{}.
\newblock \showarticletitle{Stochastic gradient descent on Riemannian
  manifolds}.
\newblock \bibinfo{journal}{\emph{IEEE Trans. Automat. Control}}
  \bibinfo{volume}{58}, \bibinfo{number}{9} (\bibinfo{year}{2013}),
  \bibinfo{pages}{2217--2229}.
\newblock


\bibitem[\protect\citeauthoryear{Buchnik and Cohen}{Buchnik and Cohen}{2018}]%
        {buchnik2018bootstrapped}
\bibfield{author}{\bibinfo{person}{Eliav Buchnik} {and} \bibinfo{person}{Edith
  Cohen}.} \bibinfo{year}{2018}\natexlab{}.
\newblock \showarticletitle{Bootstrapped graph diffusions: Exposing the power
  of nonlinearity}. In \bibinfo{booktitle}{\emph{SIGMETRICS}}.
  \bibinfo{pages}{8--10}.
\newblock


\bibitem[\protect\citeauthoryear{Cannon, Floyd, Kenyon, Parry,
  et~al\mbox{.}}{Cannon et~al\mbox{.}}{1997}]%
        {cannon1997hyperbolic}
\bibfield{author}{\bibinfo{person}{James~W Cannon}, \bibinfo{person}{William~J
  Floyd}, \bibinfo{person}{Richard Kenyon}, \bibinfo{person}{Walter~R Parry},
  {et~al\mbox{.}}} \bibinfo{year}{1997}\natexlab{}.
\newblock \showarticletitle{Hyperbolic geometry}.
\newblock \bibinfo{journal}{\emph{Flavors of geometry}}  \bibinfo{volume}{31}
  (\bibinfo{year}{1997}), \bibinfo{pages}{59--115}.
\newblock


\bibitem[\protect\citeauthoryear{Cao, Wei, Gaidon, Arechiga, and Ma}{Cao
  et~al\mbox{.}}{2019}]%
        {cao2019learning}
\bibfield{author}{\bibinfo{person}{Kaidi Cao}, \bibinfo{person}{Colin Wei},
  \bibinfo{person}{Adrien Gaidon}, \bibinfo{person}{Nikos Arechiga}, {and}
  \bibinfo{person}{Tengyu Ma}.} \bibinfo{year}{2019}\natexlab{}.
\newblock \showarticletitle{Learning imbalanced datasets with
  label-distribution-aware margin loss}. In
  \bibinfo{booktitle}{\emph{NeurIPS}}, Vol.~\bibinfo{volume}{32}.
\newblock


\bibitem[\protect\citeauthoryear{Chami, Ying, R{\'{e}}, and Leskovec}{Chami
  et~al\mbox{.}}{2019}]%
        {HGCN_ChamiYRL19}
\bibfield{author}{\bibinfo{person}{Ines Chami}, \bibinfo{person}{Zhitao Ying},
  \bibinfo{person}{Christopher R{\'{e}}}, {and} \bibinfo{person}{Jure
  Leskovec}.} \bibinfo{year}{2019}\natexlab{}.
\newblock \showarticletitle{Hyperbolic Graph Convolutional Neural Networks}. In
  \bibinfo{booktitle}{\emph{NeurIPS}}. \bibinfo{pages}{4869--4880}.
\newblock


\bibitem[\protect\citeauthoryear{Chen, Lin, Zhao, Ren, Li, Zhou, and Sun}{Chen
  et~al\mbox{.}}{2021}]%
        {chen2021topology}
\bibfield{author}{\bibinfo{person}{Deli Chen}, \bibinfo{person}{Yankai Lin},
  \bibinfo{person}{Guangxiang Zhao}, \bibinfo{person}{Xuancheng Ren},
  \bibinfo{person}{Peng Li}, \bibinfo{person}{Jie Zhou}, {and}
  \bibinfo{person}{Xu Sun}.} \bibinfo{year}{2021}\natexlab{}.
\newblock \showarticletitle{Topology-Imbalance Learning for Semi-Supervised
  Node Classification}.
\newblock \bibinfo{journal}{\emph{NeurIPS}}  \bibinfo{volume}{34}
  (\bibinfo{year}{2021}).
\newblock


\bibitem[\protect\citeauthoryear{Cui, Jia, Lin, Song, and Belongie}{Cui
  et~al\mbox{.}}{2019}]%
        {cui2019class}
\bibfield{author}{\bibinfo{person}{Yin Cui}, \bibinfo{person}{Menglin Jia},
  \bibinfo{person}{Tsung-Yi Lin}, \bibinfo{person}{Yang Song}, {and}
  \bibinfo{person}{Serge Belongie}.} \bibinfo{year}{2019}\natexlab{}.
\newblock \showarticletitle{Class-balanced loss based on effective number of
  samples}. In \bibinfo{booktitle}{\emph{CVPR}}. \bibinfo{pages}{9268--9277}.
\newblock


\bibitem[\protect\citeauthoryear{Dorogovtsev, Goltsev, and Mendes}{Dorogovtsev
  et~al\mbox{.}}{2002}]%
        {dorogovtsev2002pseudofractal}
\bibfield{author}{\bibinfo{person}{Sergey~N Dorogovtsev},
  \bibinfo{person}{Alexander~V Goltsev}, {and} \bibinfo{person}{Jos{\'e}
  Ferreira~F Mendes}.} \bibinfo{year}{2002}\natexlab{}.
\newblock \showarticletitle{Pseudofractal scale-free web}.
\newblock \bibinfo{journal}{\emph{Physical review E}} \bibinfo{volume}{65},
  \bibinfo{number}{6} (\bibinfo{year}{2002}), \bibinfo{pages}{066122}.
\newblock


\bibitem[\protect\citeauthoryear{Fu, Li, Wu, Sun, Ji, Wang, Tan, Peng, and
  Philip}{Fu et~al\mbox{.}}{2021}]%
        {fu2021ace}
\bibfield{author}{\bibinfo{person}{Xingcheng Fu}, \bibinfo{person}{Jianxin Li},
  \bibinfo{person}{Jia Wu}, \bibinfo{person}{Qingyun Sun},
  \bibinfo{person}{Cheng Ji}, \bibinfo{person}{Senzhang Wang},
  \bibinfo{person}{Jiajun Tan}, \bibinfo{person}{Hao Peng}, {and}
  \bibinfo{person}{S~Yu Philip}.} \bibinfo{year}{2021}\natexlab{}.
\newblock \showarticletitle{ACE-HGNN: Adaptive Curvature Exploration Hyperbolic
  Graph Neural Network}. In \bibinfo{booktitle}{\emph{ICDM}}. IEEE,
  \bibinfo{pages}{111--120}.
\newblock


\bibitem[\protect\citeauthoryear{Ganea, B{\'{e}}cigneul, and Hofmann}{Ganea
  et~al\mbox{.}}{2018}]%
        {Octavian2018HyperbolicNeuralNetworks}
\bibfield{author}{\bibinfo{person}{Octavian{-}Eugen Ganea},
  \bibinfo{person}{Gary B{\'{e}}cigneul}, {and} \bibinfo{person}{Thomas
  Hofmann}.} \bibinfo{year}{2018}\natexlab{}.
\newblock \showarticletitle{Hyperbolic Neural Networks}. In
  \bibinfo{booktitle}{\emph{NeurIPS}}. \bibinfo{pages}{5350--5360}.
\newblock


\bibitem[\protect\citeauthoryear{Gilmer, Schoenholz, Riley, Vinyals, and
  Dahl}{Gilmer et~al\mbox{.}}{2017}]%
        {gilmer2017neural}
\bibfield{author}{\bibinfo{person}{Justin Gilmer}, \bibinfo{person}{Samuel~S
  Schoenholz}, \bibinfo{person}{Patrick~F Riley}, \bibinfo{person}{Oriol
  Vinyals}, {and} \bibinfo{person}{George~E Dahl}.}
  \bibinfo{year}{2017}\natexlab{}.
\newblock \showarticletitle{Neural message passing for quantum chemistry}. In
  \bibinfo{booktitle}{\emph{ICML}}. \bibinfo{pages}{1263--1272}.
\newblock


\bibitem[\protect\citeauthoryear{Haixiang, Yijing, Shang, Mingyun, Yuanyue, and
  Bing}{Haixiang et~al\mbox{.}}{2017}]%
        {haixiang2017learning}
\bibfield{author}{\bibinfo{person}{Guo Haixiang}, \bibinfo{person}{Li Yijing},
  \bibinfo{person}{Jennifer Shang}, \bibinfo{person}{Gu Mingyun},
  \bibinfo{person}{Huang Yuanyue}, {and} \bibinfo{person}{Gong Bing}.}
  \bibinfo{year}{2017}\natexlab{}.
\newblock \showarticletitle{Learning from class-imbalanced data: Review of
  methods and applications}.
\newblock \bibinfo{journal}{\emph{Expert systems with applications}}
  \bibinfo{volume}{73} (\bibinfo{year}{2017}), \bibinfo{pages}{220--239}.
\newblock


\bibitem[\protect\citeauthoryear{Hamilton, Ying, and Leskovec}{Hamilton
  et~al\mbox{.}}{2017}]%
        {hamilton2017inductive}
\bibfield{author}{\bibinfo{person}{Will Hamilton}, \bibinfo{person}{Zhitao
  Ying}, {and} \bibinfo{person}{Jure Leskovec}.}
  \bibinfo{year}{2017}\natexlab{}.
\newblock \showarticletitle{Inductive representation learning on large graphs}.
  In \bibinfo{booktitle}{\emph{NeurIPS}}. \bibinfo{pages}{1024--1034}.
\newblock


\bibitem[\protect\citeauthoryear{He and Garcia}{He and Garcia}{2009}]%
        {he2009learning}
\bibfield{author}{\bibinfo{person}{Haibo He} {and} \bibinfo{person}{Edwardo~A
  Garcia}.} \bibinfo{year}{2009}\natexlab{}.
\newblock \showarticletitle{Learning from imbalanced data}.
\newblock \bibinfo{journal}{\emph{IEEE Transactions on knowledge and data
  engineering}} \bibinfo{volume}{21}, \bibinfo{number}{9}
  (\bibinfo{year}{2009}), \bibinfo{pages}{1263--1284}.
\newblock


\bibitem[\protect\citeauthoryear{Kennedy, Saniee, and Narayan}{Kennedy
  et~al\mbox{.}}{2016}]%
        {kennedy2016hyperbolicity}
\bibfield{author}{\bibinfo{person}{W~Sean Kennedy}, \bibinfo{person}{Iraj
  Saniee}, {and} \bibinfo{person}{Onuttom Narayan}.}
  \bibinfo{year}{2016}\natexlab{}.
\newblock \showarticletitle{On the hyperbolicity of large-scale networks and
  its estimation}. In \bibinfo{booktitle}{\emph{Big Data}}. IEEE,
  \bibinfo{pages}{3344--3351}.
\newblock


\bibitem[\protect\citeauthoryear{Kipf and Welling}{Kipf and Welling}{2017}]%
        {GCN}
\bibfield{author}{\bibinfo{person}{Thomas~N. Kipf} {and} \bibinfo{person}{Max
  Welling}.} \bibinfo{year}{2017}\natexlab{}.
\newblock \showarticletitle{Semi-Supervised Classification with Graph
  Convolutional Networks}. In \bibinfo{booktitle}{\emph{ICLR}}.
\newblock


\bibitem[\protect\citeauthoryear{Krioukov, Papadopoulos, Kitsak, Vahdat, and
  Bogun{\'a}}{Krioukov et~al\mbox{.}}{2010}]%
        {Krioukov2010Hyperbolic}
\bibfield{author}{\bibinfo{person}{Dmitri Krioukov},
  \bibinfo{person}{Fragkiskos Papadopoulos}, \bibinfo{person}{Maksim Kitsak},
  \bibinfo{person}{Amin Vahdat}, {and} \bibinfo{person}{Mari{\'a}n
  Bogun{\'a}}.} \bibinfo{year}{2010}\natexlab{}.
\newblock \showarticletitle{Hyperbolic geometry of complex networks}.
\newblock \bibinfo{journal}{\emph{Physical Review E}} \bibinfo{volume}{82},
  \bibinfo{number}{3} (\bibinfo{year}{2010}), \bibinfo{pages}{036106}.
\newblock


\bibitem[\protect\citeauthoryear{Li, Fu, Sun, Ji, Tan, Wu, and Peng}{Li
  et~al\mbox{.}}{2022}]%
        {li2022curvature}
\bibfield{author}{\bibinfo{person}{Jianxin Li}, \bibinfo{person}{Xingcheng Fu},
  \bibinfo{person}{Qingyun Sun}, \bibinfo{person}{Cheng Ji},
  \bibinfo{person}{Jiajun Tan}, \bibinfo{person}{Jia Wu}, {and}
  \bibinfo{person}{Hao Peng}.} \bibinfo{year}{2022}\natexlab{}.
\newblock \showarticletitle{Curvature Graph Generative Adversarial Networks}.
  In \bibinfo{booktitle}{\emph{WWW}}. \bibinfo{pages}{1528--1537}.
\newblock


\bibitem[\protect\citeauthoryear{Li, Sun, Peng, Yang, Wu, and Phillp}{Li
  et~al\mbox{.}}{2023}]%
        {li2023adaptive}
\bibfield{author}{\bibinfo{person}{Jianxin Li}, \bibinfo{person}{Qingyun Sun},
  \bibinfo{person}{Hao Peng}, \bibinfo{person}{Beining Yang},
  \bibinfo{person}{Jia Wu}, {and} \bibinfo{person}{S~Yu Phillp}.}
  \bibinfo{year}{2023}\natexlab{}.
\newblock \showarticletitle{Adaptive Subgraph Neural Network with Reinforced
  Critical Structure Mining}.
\newblock \bibinfo{journal}{\emph{IEEE Transactions on Pattern Analysis and
  Machine Intelligence}} (\bibinfo{year}{2023}).
\newblock


\bibitem[\protect\citeauthoryear{Lin, Goyal, Girshick, He, and Doll{\'a}r}{Lin
  et~al\mbox{.}}{2017}]%
        {lin2017focal}
\bibfield{author}{\bibinfo{person}{Tsung-Yi Lin}, \bibinfo{person}{Priya
  Goyal}, \bibinfo{person}{Ross Girshick}, \bibinfo{person}{Kaiming He}, {and}
  \bibinfo{person}{Piotr Doll{\'a}r}.} \bibinfo{year}{2017}\natexlab{}.
\newblock \showarticletitle{Focal loss for dense object detection}. In
  \bibinfo{booktitle}{\emph{ECCV}}. \bibinfo{pages}{2980--2988}.
\newblock


\bibitem[\protect\citeauthoryear{Nastase, Mihalcea, and Radev}{Nastase
  et~al\mbox{.}}{2015}]%
        {nastase2015survey}
\bibfield{author}{\bibinfo{person}{Vivi Nastase}, \bibinfo{person}{Rada
  Mihalcea}, {and} \bibinfo{person}{Dragomir~R Radev}.}
  \bibinfo{year}{2015}\natexlab{}.
\newblock \showarticletitle{A survey of graphs in natural language processing}.
\newblock \bibinfo{journal}{\emph{Natural Language Engineering}}
  \bibinfo{volume}{21}, \bibinfo{number}{5} (\bibinfo{year}{2015}),
  \bibinfo{pages}{665--698}.
\newblock


\bibitem[\protect\citeauthoryear{Nickel and Kiela}{Nickel and Kiela}{2017}]%
        {NickelK17Poincare}
\bibfield{author}{\bibinfo{person}{Maximilian Nickel} {and}
  \bibinfo{person}{Douwe Kiela}.} \bibinfo{year}{2017}\natexlab{}.
\newblock \showarticletitle{Poincar{\'{e}} Embeddings for Learning Hierarchical
  Representations}. In \bibinfo{booktitle}{\emph{NeurIPS}}.
  \bibinfo{pages}{6338--6347}.
\newblock


\bibitem[\protect\citeauthoryear{Ollivier}{Ollivier}{2009}]%
        {ollivier2009ricci}
\bibfield{author}{\bibinfo{person}{Yann Ollivier}.}
  \bibinfo{year}{2009}\natexlab{}.
\newblock \showarticletitle{Ricci curvature of Markov chains on metric spaces}.
\newblock \bibinfo{journal}{\emph{Journal of Functional Analysis}}
  \bibinfo{volume}{256}, \bibinfo{number}{3} (\bibinfo{year}{2009}),
  \bibinfo{pages}{810--864}.
\newblock


\bibitem[\protect\citeauthoryear{Papadopoulos, Kitsak, Serrano, Bogun{\'a}, and
  Krioukov}{Papadopoulos et~al\mbox{.}}{2012}]%
        {papadopoulos2012popularity}
\bibfield{author}{\bibinfo{person}{Fragkiskos Papadopoulos},
  \bibinfo{person}{Maksim Kitsak}, \bibinfo{person}{M~{\'A}ngeles Serrano},
  \bibinfo{person}{Mari{\'a}n Bogun{\'a}}, {and} \bibinfo{person}{Dmitri
  Krioukov}.} \bibinfo{year}{2012}\natexlab{}.
\newblock \showarticletitle{Popularity versus similarity in growing networks}.
\newblock \bibinfo{journal}{\emph{Nature}} (\bibinfo{year}{2012}),
  \bibinfo{pages}{537--540}.
\newblock


\bibitem[\protect\citeauthoryear{Park, Song, and Yang}{Park
  et~al\mbox{.}}{2021}]%
        {park2021graphens}
\bibfield{author}{\bibinfo{person}{Joonhyung Park}, \bibinfo{person}{Jaeyun
  Song}, {and} \bibinfo{person}{Eunho Yang}.} \bibinfo{year}{2021}\natexlab{}.
\newblock \showarticletitle{GraphENS: Neighbor-Aware Ego Network Synthesis for
  Class-Imbalanced Node Classification}. In \bibinfo{booktitle}{\emph{ICLR}}.
\newblock


\bibitem[\protect\citeauthoryear{Pei, Wei, Chang, Lei, and Yang}{Pei
  et~al\mbox{.}}{2020a}]%
        {peng2020GemoGCN_HG}
\bibfield{author}{\bibinfo{person}{Hongbin Pei}, \bibinfo{person}{Bingzhe Wei},
  \bibinfo{person}{Kevin~Chen{-}Chuan Chang}, \bibinfo{person}{Yu Lei}, {and}
  \bibinfo{person}{Bo Yang}.} \bibinfo{year}{2020}\natexlab{a}.
\newblock \showarticletitle{Geom-GCN: Geometric Graph Convolutional Networks}.
  In \bibinfo{booktitle}{\emph{ICLR}}. \bibinfo{publisher}{OpenReview.net}.
\newblock


\bibitem[\protect\citeauthoryear{Pei, Wei, Chang, Lei, and Yang}{Pei
  et~al\mbox{.}}{2020b}]%
        {pei2020geom}
\bibfield{author}{\bibinfo{person}{Hongbin Pei}, \bibinfo{person}{Bingzhe Wei},
  \bibinfo{person}{Kevin Chen-Chuan Chang}, \bibinfo{person}{Yu Lei}, {and}
  \bibinfo{person}{Bo Yang}.} \bibinfo{year}{2020}\natexlab{b}.
\newblock \showarticletitle{Geom-gcn: Geometric graph convolutional networks}.
  In \bibinfo{booktitle}{\emph{ICLR}}.
\newblock


\bibitem[\protect\citeauthoryear{Ravasz and Barab{\'a}si}{Ravasz and
  Barab{\'a}si}{2003}]%
        {ravasz2003hierarchical}
\bibfield{author}{\bibinfo{person}{Erzs{\'e}bet Ravasz} {and}
  \bibinfo{person}{Albert-L{\'a}szl{\'o} Barab{\'a}si}.}
  \bibinfo{year}{2003}\natexlab{}.
\newblock \showarticletitle{Hierarchical organization in complex networks}.
\newblock \bibinfo{journal}{\emph{Physical review E}} \bibinfo{volume}{67},
  \bibinfo{number}{2} (\bibinfo{year}{2003}), \bibinfo{pages}{026112}.
\newblock


\bibitem[\protect\citeauthoryear{Ren, Zeng, Yang, and Urtasun}{Ren
  et~al\mbox{.}}{2018}]%
        {ren2018learning}
\bibfield{author}{\bibinfo{person}{Mengye Ren}, \bibinfo{person}{Wenyuan Zeng},
  \bibinfo{person}{Bin Yang}, {and} \bibinfo{person}{Raquel Urtasun}.}
  \bibinfo{year}{2018}\natexlab{}.
\newblock \showarticletitle{Learning to reweight examples for robust deep
  learning}. In \bibinfo{booktitle}{\emph{ICML}}. PMLR,
  \bibinfo{pages}{4334--4343}.
\newblock


\bibitem[\protect\citeauthoryear{Rong, Huang, Xu, and Huang}{Rong
  et~al\mbox{.}}{2019}]%
        {rong2019dropedge}
\bibfield{author}{\bibinfo{person}{Yu Rong}, \bibinfo{person}{Wenbing Huang},
  \bibinfo{person}{Tingyang Xu}, {and} \bibinfo{person}{Junzhou Huang}.}
  \bibinfo{year}{2019}\natexlab{}.
\newblock \showarticletitle{Dropedge: Towards deep graph convolutional networks
  on node classification}. In \bibinfo{booktitle}{\emph{ICLR}}.
\newblock


\bibitem[\protect\citeauthoryear{Rozemberczki, Allen, and Sarkar}{Rozemberczki
  et~al\mbox{.}}{2021}]%
        {rozemberczki2021multi}
\bibfield{author}{\bibinfo{person}{Benedek Rozemberczki}, \bibinfo{person}{Carl
  Allen}, {and} \bibinfo{person}{Rik Sarkar}.} \bibinfo{year}{2021}\natexlab{}.
\newblock \showarticletitle{Multi-scale attributed node embedding}.
\newblock \bibinfo{journal}{\emph{Journal of Complex Networks}}
  \bibinfo{volume}{9}, \bibinfo{number}{2} (\bibinfo{year}{2021}),
  \bibinfo{pages}{cnab014}.
\newblock


\bibitem[\protect\citeauthoryear{Sen, Namata, Bilgic, Getoor, Galligher, and
  Eliassi-Rad}{Sen et~al\mbox{.}}{2008}]%
        {sen2008cora}
\bibfield{author}{\bibinfo{person}{Prithviraj Sen}, \bibinfo{person}{Galileo
  Namata}, \bibinfo{person}{Mustafa Bilgic}, \bibinfo{person}{Lise Getoor},
  \bibinfo{person}{Brian Galligher}, {and} \bibinfo{person}{Tina Eliassi-Rad}.}
  \bibinfo{year}{2008}\natexlab{}.
\newblock \showarticletitle{Collective classification in network data}.
\newblock \bibinfo{journal}{\emph{AI magazine}} \bibinfo{volume}{29},
  \bibinfo{number}{3} (\bibinfo{year}{2008}), \bibinfo{pages}{93--93}.
\newblock


\bibitem[\protect\citeauthoryear{Shchur, Mumme, Bojchevski, and
  G{\"u}nnemann}{Shchur et~al\mbox{.}}{2018}]%
        {shchur2018pitfalls}
\bibfield{author}{\bibinfo{person}{Oleksandr Shchur},
  \bibinfo{person}{Maximilian Mumme}, \bibinfo{person}{Aleksandar Bojchevski},
  {and} \bibinfo{person}{Stephan G{\"u}nnemann}.}
  \bibinfo{year}{2018}\natexlab{}.
\newblock \showarticletitle{Pitfalls of graph neural network evaluation}.
\newblock \bibinfo{journal}{\emph{arXiv preprint arXiv:1811.05868}}
  (\bibinfo{year}{2018}).
\newblock


\bibitem[\protect\citeauthoryear{Sia, Jonckheere, and Bogdan}{Sia
  et~al\mbox{.}}{2019}]%
        {sia2019ollivier}
\bibfield{author}{\bibinfo{person}{Jayson Sia}, \bibinfo{person}{Edmond
  Jonckheere}, {and} \bibinfo{person}{Paul Bogdan}.}
  \bibinfo{year}{2019}\natexlab{}.
\newblock \showarticletitle{Ollivier-ricci curvature-based method to community
  detection in complex networks}.
\newblock \bibinfo{journal}{\emph{Scientific reports}} \bibinfo{volume}{9},
  \bibinfo{number}{1} (\bibinfo{year}{2019}), \bibinfo{pages}{1--12}.
\newblock


\bibitem[\protect\citeauthoryear{Song, Park, and Yang}{Song
  et~al\mbox{.}}{2022}]%
        {song2022tam}
\bibfield{author}{\bibinfo{person}{Jaeyun Song}, \bibinfo{person}{Joonhyung
  Park}, {and} \bibinfo{person}{Eunho Yang}.} \bibinfo{year}{2022}\natexlab{}.
\newblock \showarticletitle{TAM: Topology-Aware Margin Loss for
  Class-Imbalanced Node Classification}. In \bibinfo{booktitle}{\emph{ICML}}.
  PMLR, \bibinfo{pages}{20369--20383}.
\newblock


\bibitem[\protect\citeauthoryear{Sun, Li, Peng, Wu, Ning, Yu, and He}{Sun
  et~al\mbox{.}}{2021}]%
        {sun2021sugar}
\bibfield{author}{\bibinfo{person}{Qingyun Sun}, \bibinfo{person}{Jianxin Li},
  \bibinfo{person}{Hao Peng}, \bibinfo{person}{Jia Wu},
  \bibinfo{person}{Yuanxing Ning}, \bibinfo{person}{Philip~S Yu}, {and}
  \bibinfo{person}{Lifang He}.} \bibinfo{year}{2021}\natexlab{}.
\newblock \showarticletitle{Sugar: Subgraph neural network with reinforcement
  pooling and self-supervised mutual information mechanism}. In
  \bibinfo{booktitle}{\emph{Proceedings of the Web Conference 2021}}.
  \bibinfo{pages}{2081--2091}.
\newblock


\bibitem[\protect\citeauthoryear{Sun, Li, Yang, Fu, Peng, and Yu}{Sun
  et~al\mbox{.}}{2023}]%
        {sun2022self}
\bibfield{author}{\bibinfo{person}{Qingyun Sun}, \bibinfo{person}{Jianxin Li},
  \bibinfo{person}{Beining Yang}, \bibinfo{person}{Xingcheng Fu},
  \bibinfo{person}{Hao Peng}, {and} \bibinfo{person}{Philip~S Yu}.}
  \bibinfo{year}{2023}\natexlab{}.
\newblock \showarticletitle{Self-organization Preserved Graph Structure
  Learning with Principle of Relevant Information}. In
  \bibinfo{booktitle}{\emph{AAAI}}.
\newblock


\bibitem[\protect\citeauthoryear{Sun, Li, Yuan, Fu, Peng, Ji, Li, and Yu}{Sun
  et~al\mbox{.}}{2022}]%
        {pastel2022}
\bibfield{author}{\bibinfo{person}{Qingyun Sun}, \bibinfo{person}{Jianxin Li},
  \bibinfo{person}{Haonan Yuan}, \bibinfo{person}{Xingcheng Fu},
  \bibinfo{person}{Hao Peng}, \bibinfo{person}{Cheng Ji}, \bibinfo{person}{Qian
  Li}, {and} \bibinfo{person}{Philip~S Yu}.} \bibinfo{year}{2022}\natexlab{}.
\newblock \showarticletitle{Position-aware Structure Learning for Graph
  Topology-imbalance by Relieving Under-reaching and Over-squashing}. In
  \bibinfo{booktitle}{\emph{CIKM}}. \bibinfo{publisher}{Association for
  Computing Machinery}, \bibinfo{pages}{1848–1857}.
\newblock


\bibitem[\protect\citeauthoryear{Sun, Wong, and Kamel}{Sun
  et~al\mbox{.}}{2009}]%
        {sun2009classification}
\bibfield{author}{\bibinfo{person}{Yanmin Sun}, \bibinfo{person}{Andrew~KC
  Wong}, {and} \bibinfo{person}{Mohamed~S Kamel}.}
  \bibinfo{year}{2009}\natexlab{}.
\newblock \showarticletitle{Classification of imbalanced data: A review}.
\newblock \bibinfo{journal}{\emph{International journal of pattern recognition
  and artificial intelligence}} \bibinfo{volume}{23}, \bibinfo{number}{04}
  (\bibinfo{year}{2009}), \bibinfo{pages}{687--719}.
\newblock


\bibitem[\protect\citeauthoryear{Tifrea, B{\'{e}}cigneul, and Ganea}{Tifrea
  et~al\mbox{.}}{2019}]%
        {PoincareGlove}
\bibfield{author}{\bibinfo{person}{Alexandru Tifrea}, \bibinfo{person}{Gary
  B{\'{e}}cigneul}, {and} \bibinfo{person}{Octavian{-}Eugen Ganea}.}
  \bibinfo{year}{2019}\natexlab{}.
\newblock \showarticletitle{Poincare Glove: Hyperbolic Word Embeddings}. In
  \bibinfo{booktitle}{\emph{ICLR}}.
\newblock


\bibitem[\protect\citeauthoryear{Topping, Di~Giovanni, Chamberlain, Dong, and
  Bronstein}{Topping et~al\mbox{.}}{2022}]%
        {topping2021understanding}
\bibfield{author}{\bibinfo{person}{Jake Topping}, \bibinfo{person}{Francesco
  Di~Giovanni}, \bibinfo{person}{Benjamin~Paul Chamberlain},
  \bibinfo{person}{Xiaowen Dong}, {and} \bibinfo{person}{Michael~M Bronstein}.}
  \bibinfo{year}{2022}\natexlab{}.
\newblock \showarticletitle{Understanding over-squashing and bottlenecks on
  graphs via curvature}. In \bibinfo{booktitle}{\emph{ICLR}}.
\newblock


\bibitem[\protect\citeauthoryear{Van~der Maaten and Hinton}{Van~der Maaten and
  Hinton}{2008}]%
        {van2008tsne}
\bibfield{author}{\bibinfo{person}{Laurens Van~der Maaten} {and}
  \bibinfo{person}{Geoffrey Hinton}.} \bibinfo{year}{2008}\natexlab{}.
\newblock \showarticletitle{Visualizing data using t-SNE.}
\newblock \bibinfo{journal}{\emph{Journal of machine learning research}}
  \bibinfo{volume}{9}, \bibinfo{number}{11} (\bibinfo{year}{2008}).
\newblock


\bibitem[\protect\citeauthoryear{V{\'a}zquez, Pastor-Satorras, and
  Vespignani}{V{\'a}zquez et~al\mbox{.}}{2002}]%
        {vazquez2002large}
\bibfield{author}{\bibinfo{person}{Alexei V{\'a}zquez},
  \bibinfo{person}{Romualdo Pastor-Satorras}, {and} \bibinfo{person}{Alessandro
  Vespignani}.} \bibinfo{year}{2002}\natexlab{}.
\newblock \showarticletitle{Large-scale topological and dynamical properties of
  the Internet}.
\newblock \bibinfo{journal}{\emph{Physical Review E}} \bibinfo{volume}{65},
  \bibinfo{number}{6} (\bibinfo{year}{2002}), \bibinfo{pages}{066130}.
\newblock


\bibitem[\protect\citeauthoryear{Velickovic, Cucurull, Casanova, Romero,
  Li{\`{o}}, and Bengio}{Velickovic et~al\mbox{.}}{2018}]%
        {GAT}
\bibfield{author}{\bibinfo{person}{Petar Velickovic}, \bibinfo{person}{Guillem
  Cucurull}, \bibinfo{person}{Arantxa Casanova}, \bibinfo{person}{Adriana
  Romero}, \bibinfo{person}{Pietro Li{\`{o}}}, {and} \bibinfo{person}{Yoshua
  Bengio}.} \bibinfo{year}{2018}\natexlab{}.
\newblock \showarticletitle{Graph Attention Networks}. In
  \bibinfo{booktitle}{\emph{ICLR}}.
\newblock


\bibitem[\protect\citeauthoryear{Wang, Aggarwal, and Derr}{Wang
  et~al\mbox{.}}{2021}]%
        {wang2021distance}
\bibfield{author}{\bibinfo{person}{Yu Wang}, \bibinfo{person}{Charu Aggarwal},
  {and} \bibinfo{person}{Tyler Derr}.} \bibinfo{year}{2021}\natexlab{}.
\newblock \showarticletitle{Distance-wise Prototypical Graph Neural Network in
  Node Imbalance Classification}.
\newblock \bibinfo{journal}{\emph{arXiv preprint arXiv:2110.12035}}
  (\bibinfo{year}{2021}).
\newblock


\bibitem[\protect\citeauthoryear{Ye, Liu, Ma, Gao, and Chen}{Ye
  et~al\mbox{.}}{2019}]%
        {ye2019curvature}
\bibfield{author}{\bibinfo{person}{Ze Ye}, \bibinfo{person}{Kin~Sum Liu},
  \bibinfo{person}{Tengfei Ma}, \bibinfo{person}{Jie Gao}, {and}
  \bibinfo{person}{Chao Chen}.} \bibinfo{year}{2019}\natexlab{}.
\newblock \showarticletitle{Curvature graph network}. In
  \bibinfo{booktitle}{\emph{ICLR}}.
\newblock


\bibitem[\protect\citeauthoryear{Yu, Liu, Wei, Zhou, Nakata, Gudovskiy, Okuno,
  Li, Keutzer, and Zhang}{Yu et~al\mbox{.}}{2022}]%
        {yu2022cross}
\bibfield{author}{\bibinfo{person}{Jinze Yu}, \bibinfo{person}{Jiaming Liu},
  \bibinfo{person}{Xiaobao Wei}, \bibinfo{person}{Haoyi Zhou},
  \bibinfo{person}{Yohei Nakata}, \bibinfo{person}{Denis Gudovskiy},
  \bibinfo{person}{Tomoyuki Okuno}, \bibinfo{person}{Jianxin Li},
  \bibinfo{person}{Kurt Keutzer}, {and} \bibinfo{person}{Shanghang Zhang}.}
  \bibinfo{year}{2022}\natexlab{}.
\newblock \showarticletitle{Cross-Domain Object Detection with Mean-Teacher
  Transformer}. In \bibinfo{booktitle}{\emph{ECCV}}.
\newblock


\bibitem[\protect\citeauthoryear{Zhang, Cui, and Zhu}{Zhang
  et~al\mbox{.}}{2020}]%
        {zhang2020deep}
\bibfield{author}{\bibinfo{person}{Ziwei Zhang}, \bibinfo{person}{Peng Cui},
  {and} \bibinfo{person}{Wenwu Zhu}.} \bibinfo{year}{2020}\natexlab{}.
\newblock \showarticletitle{Deep learning on graphs: A survey}.
\newblock \bibinfo{journal}{\emph{IEEE Transactions on Knowledge and Data
  Engineering}} (\bibinfo{year}{2020}).
\newblock


\end{thebibliography}
